\documentclass[lettersize,journal]{IEEEtran}
\usepackage{amsmath,amsfonts}
\usepackage{algorithmic}
\usepackage{algorithm}
\usepackage{array}
\usepackage[caption=false,font=normalsize,labelfont=sf,textfont=sf]{subfig}
\usepackage{textcomp}
\usepackage{stfloats}
\usepackage{url}
\usepackage{verbatim}
\usepackage{graphicx}
\usepackage{cite}
\usepackage{xspace}
\usepackage{float}
\usepackage{xcolor}
\usepackage{dsfont}
\usepackage{rotating}
\usepackage{listings}
\usepackage{mathtools}
\definecolor{codegreen}{rgb}{0,0.6,0}
\definecolor{codegray}{rgb}{0.5,0.5,0.5}
\definecolor{codepurple}{rgb}{0.58,0,0.82}
\definecolor{backcolour}{rgb}{0.95,0.95,0.92}

\lstdefinestyle{mystyle}{
    backgroundcolor=\color{backcolour},   
    commentstyle=\color{codegreen},
    keywordstyle=\color{magenta},
    numberstyle=\tiny\color{codegray},
    stringstyle=\color{codepurple},
    basicstyle=\ttfamily\footnotesize,
    breakatwhitespace=false,         
    breaklines=true,                 
    captionpos=b,                    
    keepspaces=true,                 
    numbers=left,                    
    numbersep=5pt,                  
    showspaces=false,                
    showstringspaces=false,
    showtabs=false,                  
    tabsize=2
}

\usepackage{amsthm}
\usepackage{amsmath}
\renewcommand{\epsilon}{\varepsilon}

\usepackage[utf8]{inputenc}
\usepackage{verbatim,graphicx}
\usepackage{hyperref}
\usepackage{soul}
\usepackage{url,placeins}

\usepackage{amsmath,amsfonts,bm}

\def\eqref#1{equation~\ref{#1}}

\def\1{\bm{1}}

\def\vx{{\bm{x}}}

\DeclareMathAlphabet{\mathsfit}{\encodingdefault}{\sfdefault}{m}{sl}
\SetMathAlphabet{\mathsfit}{bold}{\encodingdefault}{\sfdefault}{bx}{n}

\begin{document}

\title{Optimizing with Low Budgets: a Comparison on the Black-box Optimization Benchmarking Suite and OpenAI Gym}

\author{Elena Raponi, Nathanaël Carraz Rakotonirina, J\'er\'emy Rapin, Carola Doerr, Olivier Teytaud
\thanks{E. Raponi (the corresponding author) is with LIACS, Leiden University, Leiden, The Netherlands (e-mail: e.raponi@liacs.leidenuniv.nl). At the time of writing this article, she was initially affiliated with Meta AI Research in Paris, France, and later with the Technical University of Munich in Munich, Germany, and LIP6, CNRS, Sorbonne Université in Paris, France.  
N. C. Rakotonirina is with Universitat Pompeu Fabra, Barcelona, Spain (e-mail: nathanael.rakotonirina@upf.edu). At the time of writing this article, he was affiliated with LIMA, Université d'Antananarivo, Madagascar.  
J. Rapin and O. Teytaud are with Meta AI Research, Paris, France (e-mail: {jrapin, oteytaud}@fb.com). 
C. Doerr is with LIP6, CNRS, Sorbonne Universit\'e, Paris, France (e-mail: carola.doerr@lip6.fr).}%
}

\markboth{}{Raponi, Rakotonirina, Rapin, Doerr, Teytaud}%

\maketitle

\begin{abstract}
The growing ubiquity of machine learning (ML) has led it to enter various areas of computer science, including black-box optimization (BBO). Recent research is particularly concerned with Bayesian optimization (BO). BO-based algorithms are popular in the ML community, as they are used for hyperparameter optimization and more generally for algorithm configuration. However, their efficiency decreases as the dimensionality of the problem and the budget of evaluations increase. Meanwhile, derivative-free optimization methods have evolved independently in the optimization community. Therefore, we urge to understand whether cross-fertilization is possible between the two communities, ML and BBO, i.e., whether algorithms that are heavily used in ML also work well in BBO and vice versa. Comparative experiments often involve rather small benchmarks and show visible problems in the experimental setup, such as poor initialization of baselines, overfitting due to problem-specific setting of hyperparameters, and low statistical significance.

With this paper, we update and extend a comparative study presented by Hutter et al. in 2013. We compare BBO tools for ML with more classical heuristics, first on the well-known BBOB benchmark suite from the COCO environment and then on Direct Policy Search for OpenAI Gym, a reinforcement learning benchmark. Our results confirm that BO-based optimizers perform well on both benchmarks when budgets are limited, albeit with a higher computational cost, while they are often outperformed by algorithms from other families when the evaluation budget becomes larger. We also show that some algorithms from the BBO community perform surprisingly well on ML tasks.
\end{abstract}


\section{Introduction} 

Black-Box Optimization (BBO) is an affirmed and rapidly growing field of optimization and a topic of critical importance in many application areas including complex systems engineering, energy and environment, materials design, drug discovery, chemical process synthesis, and computational biology \cite{bajaj_black-box_2021}. 
As in other classical optimization contexts, BBO assumes that we are facing an objective function $f$ for which we aim to provide a solution $x$ with $f(x)$ as good as possible using as little computational effort as possible. The key distinguishing property of BBO is that the algorithms learn about the problem instance $f$ only by querying the quality $f(x)$ of possible solution candidates $x$. This may be because we indeed lack an explicit representation of $f$ (e.g., if simulations or experiments are required for assessing the quality of a possible solution) or when we lack efficient approaches to make use of the instance information (e.g., many real-world scheduling and routing problems are solved with heuristic approaches). Simplifying the complexity measurement, in BBO we typically only account for the number of function evaluations, and hence aim for identifying high-quality solutions $x$ using as few function evaluations as possible. 

Because of its high practical relevance, people with many different backgrounds are drawn into BBO, leading to a multitude of approaches in the area, ranging from simple heuristics such as local search to local/global modeling approaches. To understand the strengths and weaknesses of these different methods, fair performance comparisons are needed. Several platforms and benchmark suites (i.e., collections of benchmark problems) address this empirical comparison. Some examples are the Black-Box Optimization Benchmarking (BBOB) collection of the COCO environment~\cite{hansen_coco_2021}, Large-Scale Global Optimization (LSGO) \cite{tang_benchmark_2010}, Nevergrad \cite{nevergrad}, 
Pseudo-Boolean Optimization (PBO) \cite{IOHprofiler}, 
and Machine Learning and Data Analysis (MLDA)~\cite{rapin_exploring_2019}. Apart from these more general benchmarking suites, there are also problem collections for evaluating and comparing algorithms for specific BBO tasks such as algorithm configuration and selection~\cite{ASlib,AClib}, neural architecture search~\cite{NAShutte,NASbench}, and expensive global optimization~\cite{expobench}; see~\cite[Section~3]{BartzSurveyArxiv} for a more exhaustive summary. 

An approach commonly used for expensive optimization problems (for which the available number of function evaluations can be very small) uses surrogates to approximate the problem instance $f$, with the idea that the approximation $\hat{f}$ can be used to locate interesting solution candidates without requiring evaluations of the true problem $f$.  
Recently, Machine Learning (ML) has gone down this road and proposed several tools for BBO along these lines.
In particular, a large field of research is Bayesian Optimization (BO), which is based on the Efficient Global Optimization (EGO) algorithm \cite{jones_efficient_1998}.
Despite its success, BO is stated to be limited to less than 15 parameters \cite{nayebi_framework_2019, wang_bayesian_2016-2} and a few thousand evaluations \cite{wang_batched_2018} according to the literature.
To overcome this issue, recent research started to explore space partitioning and local modeling. In fact, learning a classifier that locates the samples on a promising subregion of the domain with high probability might be more effective than learning a regressor on the whole domain.
Among other partitioning strategies, the Latent Action Monte Carlo Tree Search (LA-MCTS) \cite{lamctsNeurips} recursively learns space partition in a hierarchical manner using Monte Carlo Tree Search (MCTS) \cite{coulom06}. 
Therefore, in addition to BO, MCTS has also been adapted from control and games to BBO \cite{munosmctsbbo,lamctsNeurips}. 
However, both the BO and MCTS tools for BBO have rarely been compared to existing BBO methods in a systematic and satisfactory manner. 
For example, the comparisons in the LA-MCTS paper~\cite{lamctsNeurips}  depend heavily on poor initialization of competitors, and the baselines used in the paper are not made available in the provided code, while the comparisons in~\cite{TurnerEMKLXG20} consider only the simple $(1+1)$ sampling method and not the more sophisticated (and often better performing) BBO methods provided by the Nevergrad platform, although they refer to the comparison as `Nevergrad'. 
However, we note some efforts to make neutral and comprehensive comparisons. 
Extensive comparisons in Nevergrad \cite{nevergrad} tend to favor more classical methods such as tools from mathematical programming like Cobyla \cite{cobyla} and others \cite{bobyqa,powell} or evolution strategies \cite{Beyer:bookES} like CMA-ES \cite{cma}, possibly equipped with surrogate models \cite{auger_local_2005,metadyna,smds}. 
Hutter et al. \cite{bbobsmac} ran SMAC-BBOB, a well-known BO framework, on the BBOB benchmark suite and got positive results for BO when the budget of admissible function evaluations is fairly small. 
However, their investigation on expensive black-box functions only assesses SMAC against CMA-ES, yielding a rather limited benchmarking study.

Overall, there is widespread utilization of black-box optimization algorithms in the ML field, yet there exists a deficiency in conducting comprehensive comparisons between the algorithms favored in ML domains and those favored by evolutionary computation researchers. Recent papers\cite{rlgoogle} raised doubts on the reproducibility of some results in the ML community.
In this work, we update and extend the comparison made in~\cite{bbobsmac} by adding several state-of-the-art solvers and by comparing not only on the BBOB benchmark suite but also -- as it is closer to ML -- on direct policy search for OpenAI Gym problems~\cite{brockman_openai_2016}. 
We evaluate the performance of various BBO algorithms, considering solvers commonly associated with the ML community as well as more classical heuristics, and testing over a range of budgets that allow us to draw fair and unbiased conclusions.

\section{Black-box Optimization Algorithms}
\label{sec:algos}

We briefly summarize the algorithms included in our comparison, along with a brief description of two main classes of interest, BO and LA-MCTS. 

\subsection{Bayesian Optimization}
\label{sec:BO}

BO~\cite{jones_efficient_1998,mockus_bayesian_2012} is a sequential design strategy targeting global optimization of black-box functions that do not assume any functional forms. It is particularly advantageous for problems where the objective function is difficult to evaluate, is a black box with some known structure, relies upon less than 20 dimensions, and where no information about sensitivity and derivatives is available. Since the objective function does not have an explicit mathematical formulation, BO treats it as a random function and places a prior over it. 
A Kriging model, also known as \emph{Gaussian Process Regression} (GPR), can be used as a prior probability distribution over functions in BO.

BO starts with sampling an initial Design of Experiments (DoE) of size $n_0$: $\mathbf{X}=[\mathbf{x}^{(1)},\mathbf{x}^{(2)}, \ldots, \mathbf{x}^{(n_0)}]^\top$, where the sample $i$ is denoted as $\mathbf{x}^{(i)} = (x^{(i)}_1, ..., x^{(i)}_D) \in \mathcal{X} \subset \mathbb{R}^D$~\cite{santner_design_2003}. 
The corresponding objective function values are denoted as $\mathbf{y}=(f(\mathbf{x}^{(1)}),f(\mathbf{x}^{(2)}),\ldots,f(\mathbf{x}^{(n_0)}))^\top$. 
Conventionally, a centered Gaussian process prior is assumed on the objective function: $f \sim gp(0, K(\cdot, \cdot))$, where $K\colon \mathcal{X} \times \mathcal{X} \rightarrow \mathbb{R}$ is a positive definite function -- also known as ~\textit{kernel function} -- which computes the autocovariance of the process.

Often, a Gaussian likelihood is taken, leading to a conjugate posterior process~\cite{rasmussen_gaussian_2006}, i.e., $f \;|\; \mathbf{y} \sim gp(\hat{f}(\cdot), K'(\cdot, \cdot))$, where $\hat{f}$ and $K'$ are the posterior mean and covariance function, respectively.

On an unknown point $\mathbf{x}$, $\hat{f}(\mathbf{x})$ yields the maximum \emph{a posteriori} estimate of $f(\mathbf{x})$ whereas $\hat{s}^2(\mathbf{x}) \coloneqq K'(\mathbf{x}, \mathbf{x})$ quantifies the uncertainty of this estimation. 
The posterior process is, again, a GPR. Based on the posterior process, promising points are identified via the so-called \textit{infill-criterion}, i.e., by optimizing an \textit{acquisition function} that balances $\hat{f}$ with $\hat{s}^2$ (exploitation vs. exploration). A variety of infill-criteria has been proposed in the literature, e.g., Probability of Improvement~\cite{forrester_engineering_2008,mockus_bayesian_2012}, Expected Improvement~\cite{forrester_engineering_2008}, and the Upper Confidence Bound~\cite{lai1985asymptotically, srinivas_gaussian_2012}.
When a new candidate point is selected by the infill criterion, it is evaluated and added to the BO data set, which is used to update the GPR posterior.
This process is repeated until a stopping condition is met, i.e., a good enough result is located or the computational budget is exhausted.

\subsection{Monte Carlo Tree Search}
\label{sec:LAMCTS}

Monte Carlo Tree Search (MCTS) \cite{coulom06} is a solver that migrated from trees of bandits for games and control, including alpha-zero \cite{alphago,alphazero}, to applications in BBO \cite{munos_bandits_2014-1,lamctsNeurips}.
Its evolved version, LA-MCTS \cite{lamctsNeurips}, progressively learns and generalizes promising regions in the problem space by recursively partitioning so that solvers like BO can access these regions to improve their performance. 
At any iteration~$t$ of the algorithm, a training dataset $\mathbf{D}_t = (\mathbf{X},\mathbf{Y})$ of all the points evaluated so far is available.
A tree node~$A$ represents a subregion of the search space $\Omega_A$. Therefore, $\mathbf{D}_t \cap \Omega_A$ is the set of all the samples falling in the subregion $\Omega_A$. 
MCTS uses the Monte Carlo simulation to accumulate value estimates that lead to highly rewarding trajectories in the search tree. In other words, MCTS pays more attention to promising nodes (i.e., subregions of the search space), in order to avoid the need to brute force all possibilities.
This is done by using an Upper Confidence Bound (UCB) policy.
More specifically, each node has an associated UCB value and, during selection, the child node with the highest UCB value is considered.
The statistics used to compute the UCB are (1) $n_A$, which is the number of samples in $\mathbf{D}_t \cap \Omega_A$, and (2) the node value $v_A:=(\sum_{\vx_i \in \mathbf{D}_t} \cap \Omega_A f(\vx_i))/n_A$. 

Therefore, in LA-MCTS, which is the MCTS-based optimizer that we include in the comparisons of this study, promising regions are found by recursively partitioning the search space based on latent actions. 
In one iteration, LA-MCTS starts building the tree by partitioning and then selects a region based on UCB. Finally, sampling is performed in the selected region using BO.
In this way, BO avoids over-exploring the search space, and its performance improves, especially for high-dimensional problems.

\subsection{Selection of BO-based Algorithms} 
From the large collection of existing BO-based solvers, we have selected the following ones for our empirical comparison:

\begin{itemize}
    \item BO: the Bayesian Optimization algorithm~\cite{bo} implemented in Nevergrad. The python class is a wrapper over the bayes\_opt package \cite{bopackage}.
        
    \item LA-MCTS \cite{lamctsNeurips}: as described above, LA-MTCS is an MCTS-based derivative-free meta-solver that recursively learns space partition in a hierarchical manner. Sampling is then performed in the selected region using BO.
    
    \item Turbo~\cite{turbo}: a trust-region-inspired algorithm using Thompson sampling rather than the optimization of an acquisition function to find new candidate solutions in each subregion. Turbo20 denotes the multi-trust-regions counterpart of Turbo.

    \item AX~\cite{ax}: a modular BO framework that uses BoTorch primitives for optimization over continuous spaces. It automates the selection of optimization routines, reducing the amount of fine-tuning required.
    
    \item SMAC~\cite{smac}: a sequential model-based algorithm for algorithm configuration to optimize the parameters of arbitrary algorithms. It scales well to high dimensions and is particularly suitable for hyperparameter optimization of ML algorithms. The main core consists of BO. SMAC2 refers to SMAC-HPO, i.e., SMAC with hyperparameter optimization. 
    
    \item HyperOpt~\cite{HyperOpt}: library for serial and parallel hyperparameter optimization, designed to accommodate BO algorithms based on Gaussian processes and regression trees. We use the version based on Parzen estimates. 
    
    \item Optuna~\cite{OptunaKDD}: automatic hyperparameter optimization software framework which uses state-of-the-art algorithms for sampling hyperparameters and pruning unpromising trials. By default, Optuna implements a BO algorithm (Tree-structured Parzen Estimator).

\end{itemize}

\subsection{Classical Black-box Optimization Algorithms}
\label{sec:two}

As baselines commonly used in the broader BBO context we consider \textbf{CMA}, which stands for CMA-ES, a well-known evolution strategy~\cite{cma}, \textbf{Cobyla}, a tool from mathematical programming~\cite{cobyla}, particle swarm optimization (PSO~\cite{psoreference}), and~\textbf{NGOpt} from Nevergrad \cite{nevergrad}, a wizard that combines many classical algorithms in various ways~\cite{MeunierRWRRTMD22}. In the use cases considered in this work (sequential, low-dimensional, noise-free problems), NGOpt mainly uses CMA, Cobyla, and (1+1)-type sampling equipped with metamodels. 

We also include DefaultCMA\cite{cma}, a version of CMA without the BBOB-specific initialization used for the experiments of CMA on BBOB \cite{bbob}. We do this to show that ad hoc initialization of CMA-ES does not lead to significant improvement over DefaultCMA.

\section{Benchmark Problems}
\label{modif}

Extensive comparisons have already been proposed in Nevergrad~\cite{MeunierRWRRTMD22}, focusing on reproducibility, real-world, and different problem sizes. 
We propose here additional experiments in the well-known BBOB framework \cite{bbob}, chosen for its simplicity/canonicity, and for OpenAI Gym\footnote{\url{https://www.gymlibrary.dev/}}, commonly used in the reinforcement learning (RL) environment \cite{gym1,gym2,gymcopter,gymf,gymanm,radiogym,green_impacts_2018, gymrepro,gym5g}. 

\textbf{BBOB:} The BBOB collection contains 24 functions, with known difficulty (e.g., non-separability, ill-conditioning, different levels of multimodality, adequate or weak global structure, etc.).
Although all functions are defined and can be evaluated over $\mathbb{R}^D$, the default search domain is $[-5, 5]^D$; that is, in contrast to Nevergrad's experiments in \cite{nevergrad,dash}, the BBOB suite has a focus on bounded domains. 

BBOB offers a possibility to randomize both the position of the optimal solution (that is, test functions are randomly shifted in the domain) and the function value of the optimum (test functions are randomly shifted in the co-domain). To reduce bias with respect to problem encoding, we run the algorithms on 15 randomly chosen instances per test function and dimension. 
We consider six different dimensions (2, 3, 5, 10, 20, and 40). 

As suggested in \cite{bbobsmac}, we focus on a relatively small budget of $10D$ or $100D$ function evaluations (the default setting for BBOB experiments is $1000D$). When a method crashes, we rerun it with the remaining budget. 

For the BBOB experiments, our key performance measure is the empirical cumulative distribution function (ECDF) of the runtimes needed to reach the optimal objective value with a given precision $\Delta t$, i.e., the runtime depends on a given target function value ($f_t = f_\text{opt} + \Delta t$) and is computed across all runs of an optimizer on a function, as the total number of function evaluations before reaching $f_t$ divided by the number of trials that actually reached $f_t$. Informally, the ECDF shows the proportion of problems solved within a given budget, with the budget indicated on the x-axis.

\textbf{OpenAI Gym with Direct Policy Search:} Firstly, we work on a multi-deterministic Open AI Gym with tiny neural nets. Indeed, we analyze a small version (small number of neurons, low budget) to fit the low-budget context of the present work.  
\emph{Multi-deterministic} here means that for a given algorithm, a different seed is drawn at random for each repetition, and this list of seeds is used for all algorithms. There are different seeds so that we avoid overfitting, and the list of seeds is the same over the different algorithms so that we have statistical pairing.
This means that parameters are not tuned based on specific landscapes (e.g., bounded/unbounded domains, prescribed optimal regions, and multimodality of the objective function). This is somewhat analogous to random perturbation of the optimum in classical BBO benchmarks in the sense that the objective function is deterministic but drawn randomly. 

Within the gym library, we select a few environments to compare algorithms based on (1) average loss (i.e. average negated reward) 
and (2) average winning rates (i.e., the frequencies at which one method outperforms another): 
MountainCar-v0 ($D = 12$), 
Pendulum-v0 ($D=15$),
GuessingGame-v0 ($D = 24$), 
HotterColer-v0 ($D = 24$),
CartPole-v0 ($D = 28$).
CartPole-v1 ($D = 28$), 
NChain-v0 ($D = 40$), and 
MountainCarContinuous-v0 ($D = 8$). 
We chose these problems with $D < 50$ because they are sufficiently challenging and are not too hard in our context of tiny networks and minimal budget, i.e., not all algorithms perform equally. Here we use a \emph{neural factor} (i.e., the scaling coefficient used in the benchmarking suite to choose the size of neural networks) of 1. The dimension, i.e., the total number of weights, is a consequence of the number of neurons, which in turn is based on the scaling factor (the number of neurons per hidden layer is the neural factor multiplied by the input dimension).

In a second set of experiments we include problems with dimensions $D \leq 264$ for larger networks defined by setting the neural factor to $3$, which parameterizes the size of the hidden layer. This bound on the dimension and the different neural factor lead to a different subset of OpenAI Gym problems: 
LunarLanderContinuous-v2,
Blackjack-v0,
Pendulum-v0,
HotterColder-v0,
MountainCarContinuous-v0,
CartPole-v0,
Acrobot-v1,
NChain-v0,
GuessingGame-v0,
CartPole-v1.

Winning percentage rates are evaluated at fixed budgets of 25, 50, 100, 200, and 400 function evaluations for both settings (neural factor $1$ and $3$), while bigger budgets are also considered for the bigger networks (see Sections IV and V in the Supplementary Material).

\section{Results of the Empirical Comparison} 
\begin{figure*}[b]
\centering
\subfloat[]{\includegraphics[ clip, width=.32\textwidth]{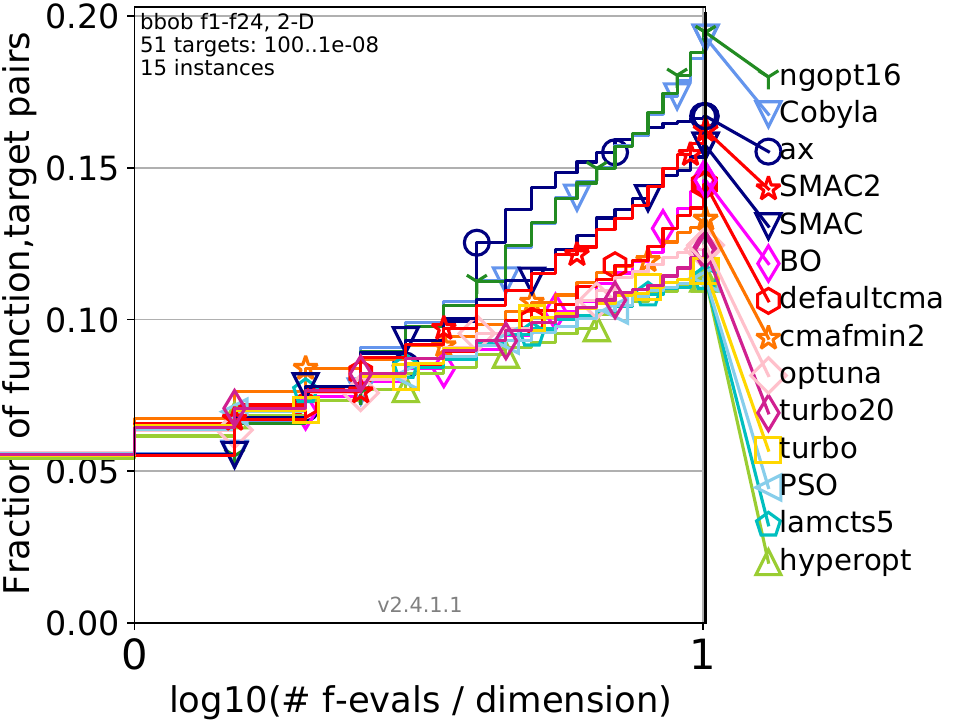}}
\subfloat[]{\includegraphics[ clip, width=.32\textwidth]{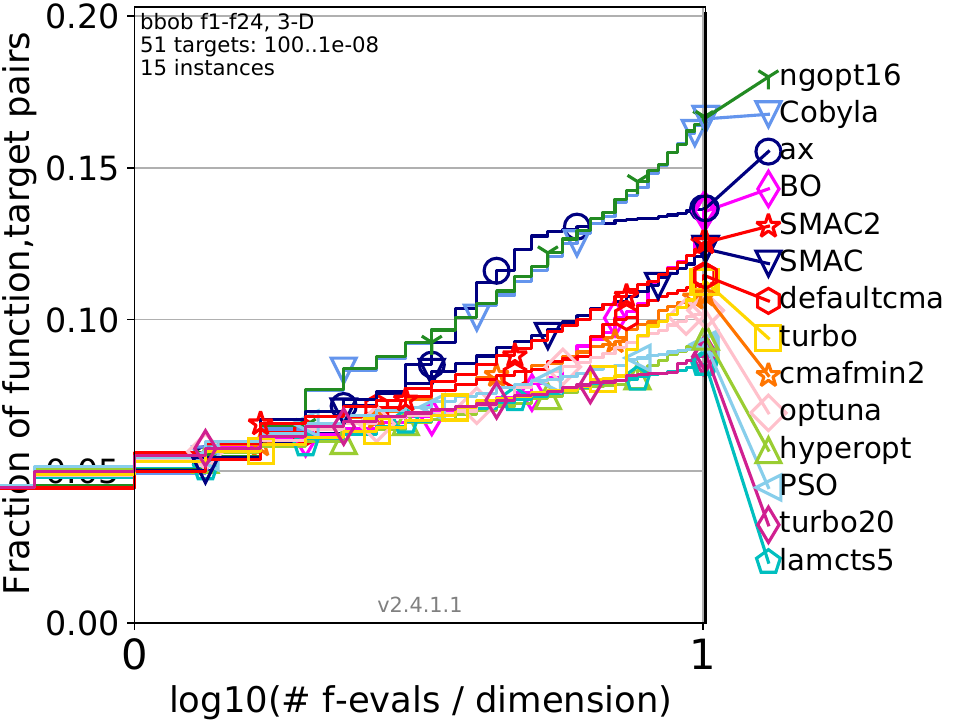}}
\subfloat[]{\includegraphics[ clip,width=.32\textwidth]{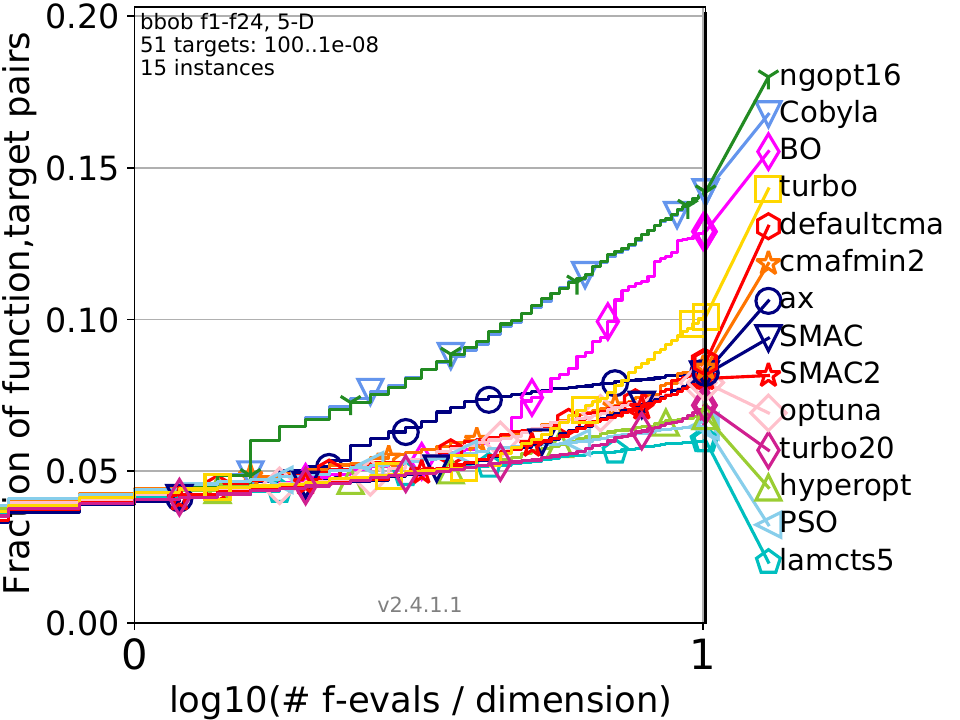}}
\\
\subfloat[]{\includegraphics[ clip,width=.32\textwidth]{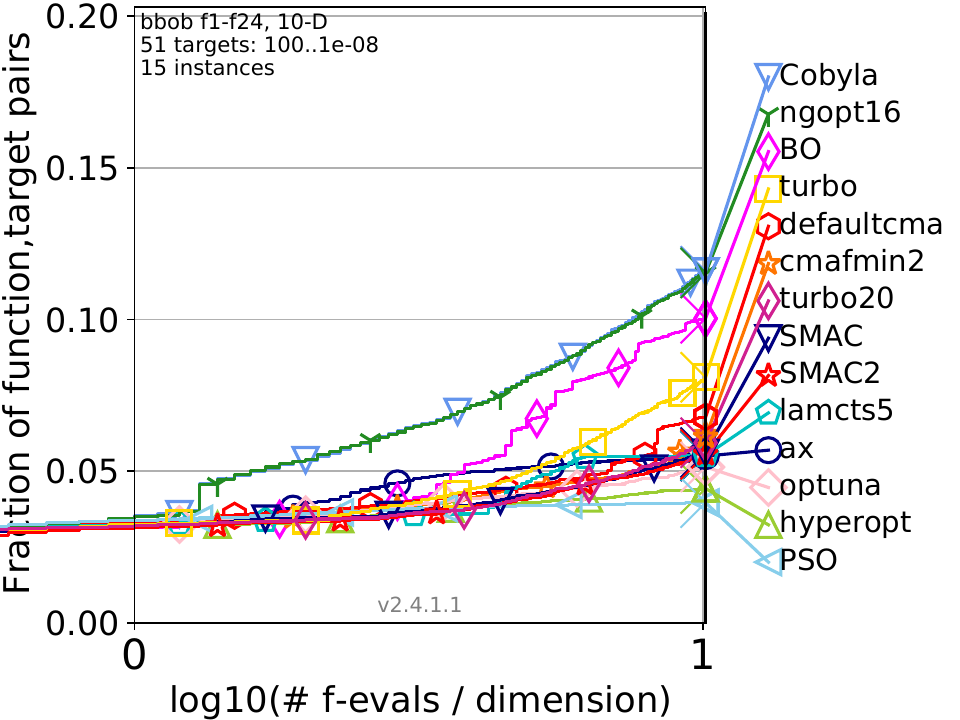}}
\subfloat[]{\includegraphics[ clip,width=.32\textwidth]{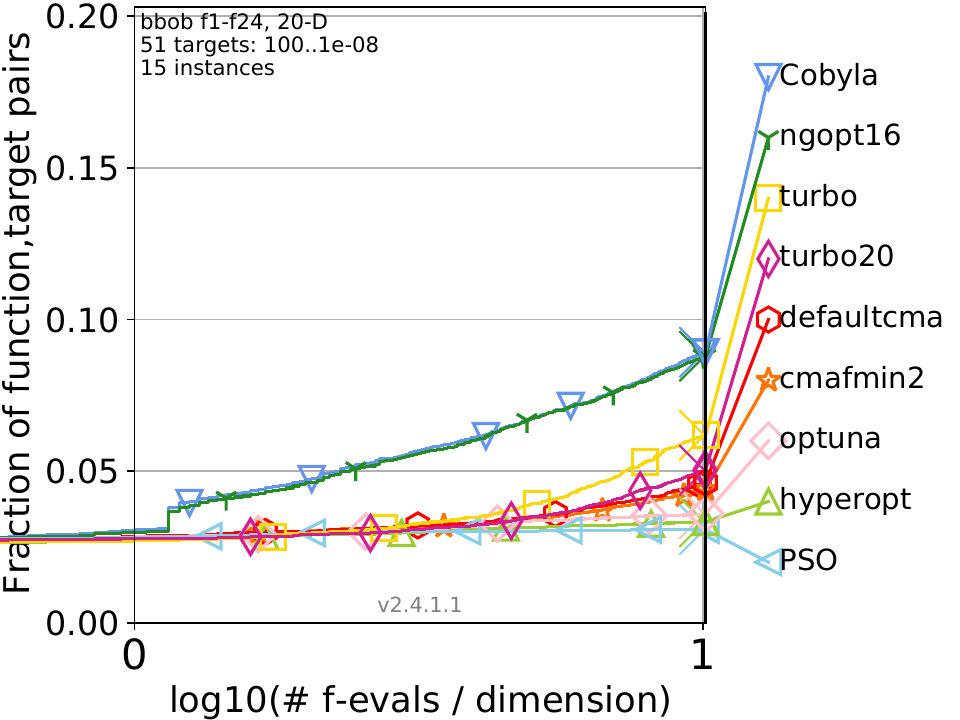}}
\subfloat[]{\includegraphics[ clip,width=.32\textwidth]{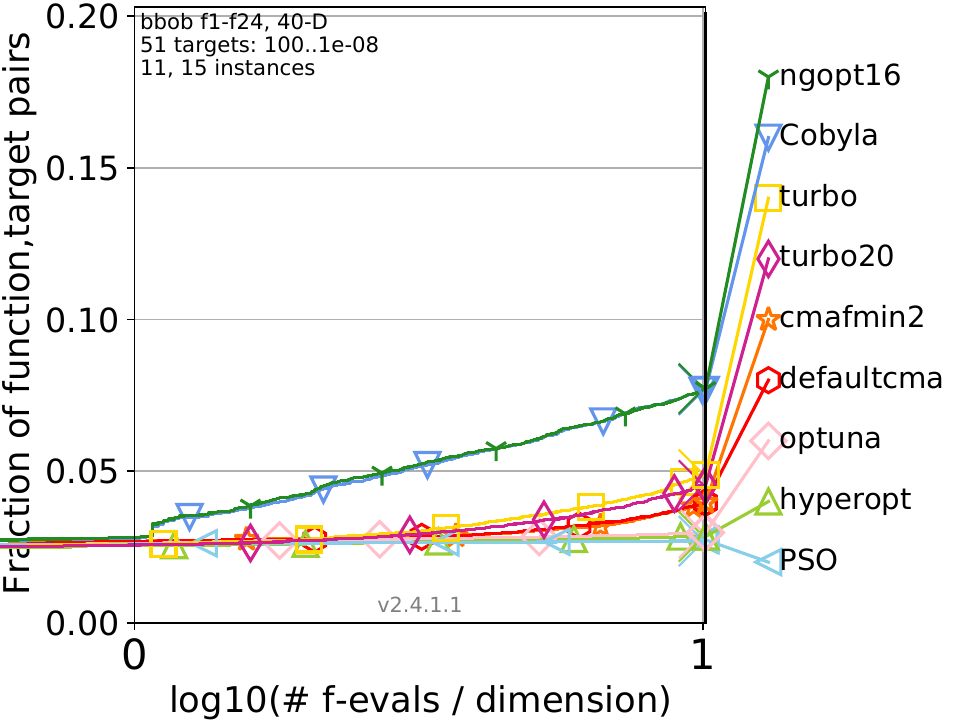}}
\caption{\label{b10d} 
Illustration of the ECDF of runtimes on the BBOB functions f1--f24, using 51 targets uniformly spaced on a log scale between $1e-8$ and $100$. Plots are shown for dimension D equal to (a) 2, (b) 3, (c) 5, (d) 10, (e) 20, and (f) 40 and budget $10D$. X-axis: budget/dimension in log-scale. Y-axis: frequency of solving at the requested precision. 
Overall, Cobyla and NGOpt16 (which heavily relies on Cobyla) perform best in these examples.
}
\label{fig_sim}
\end{figure*}

\begin{figure*}[!h]
\centering
\subfloat[]{\includegraphics[ clip,width=.32\textwidth]{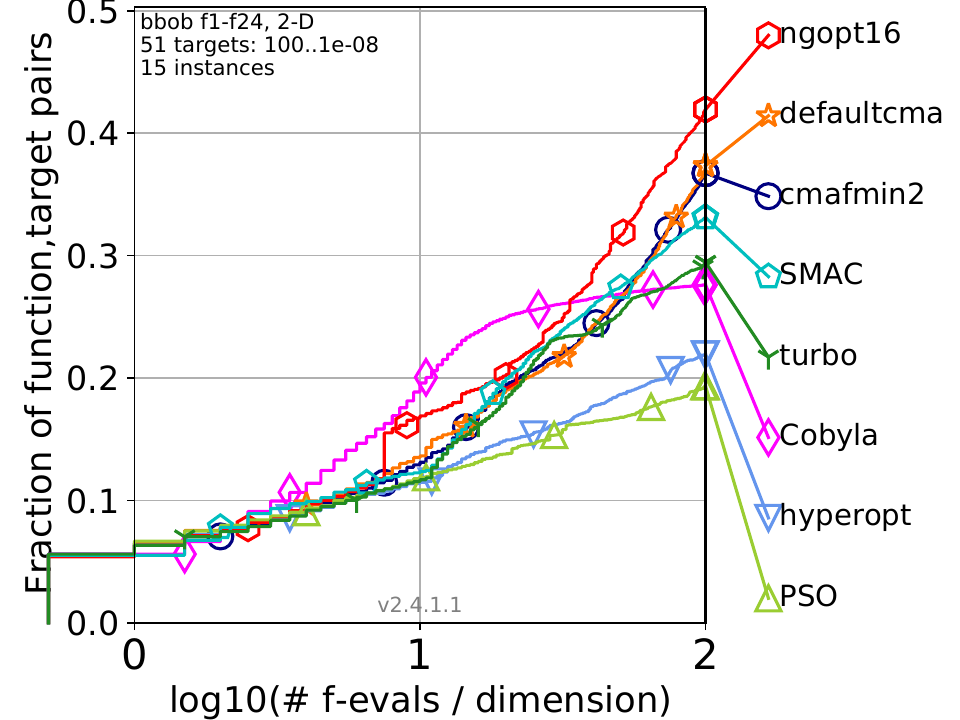}}\qquad
\subfloat[]{\includegraphics[ clip,width=.32\textwidth]{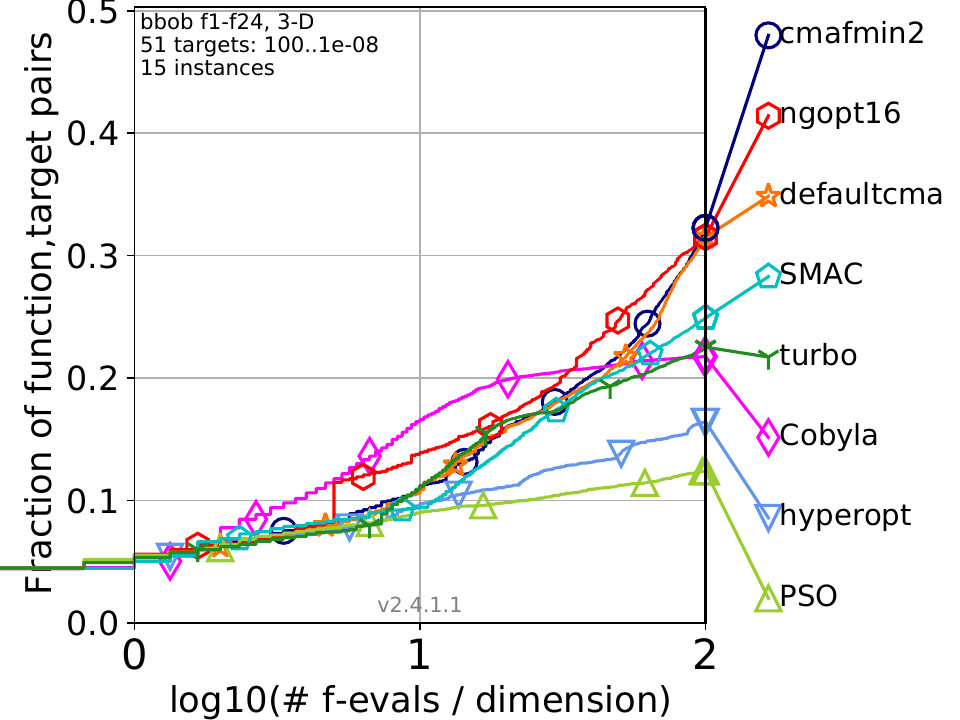}}\\
\subfloat[]{\includegraphics[ clip,width=.32\textwidth]{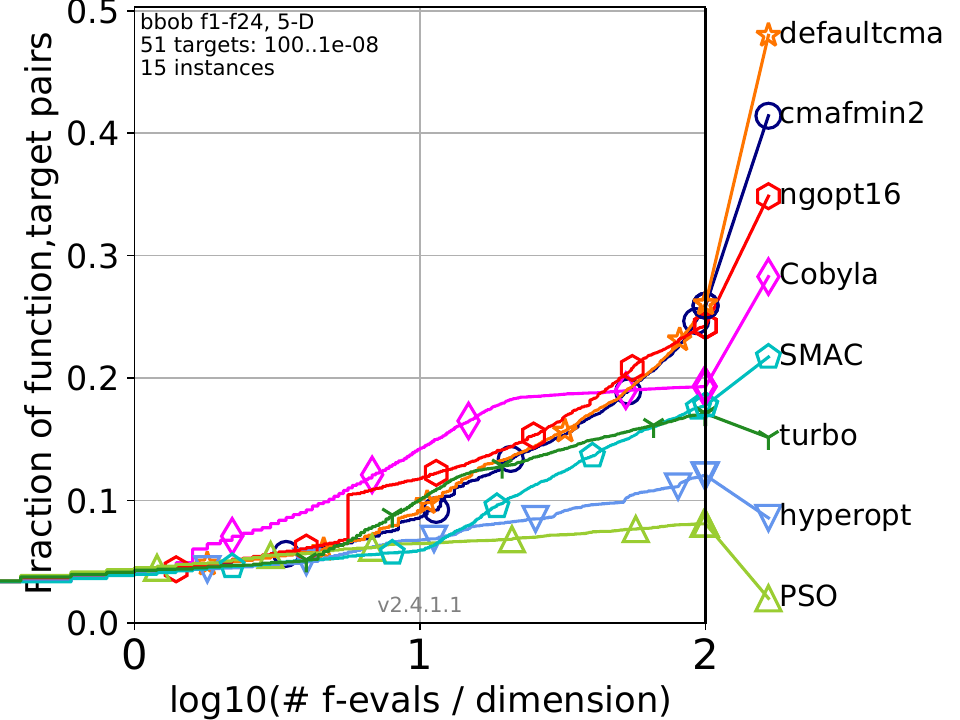}}\qquad
\subfloat[]{\includegraphics[ clip,width=.32\textwidth]{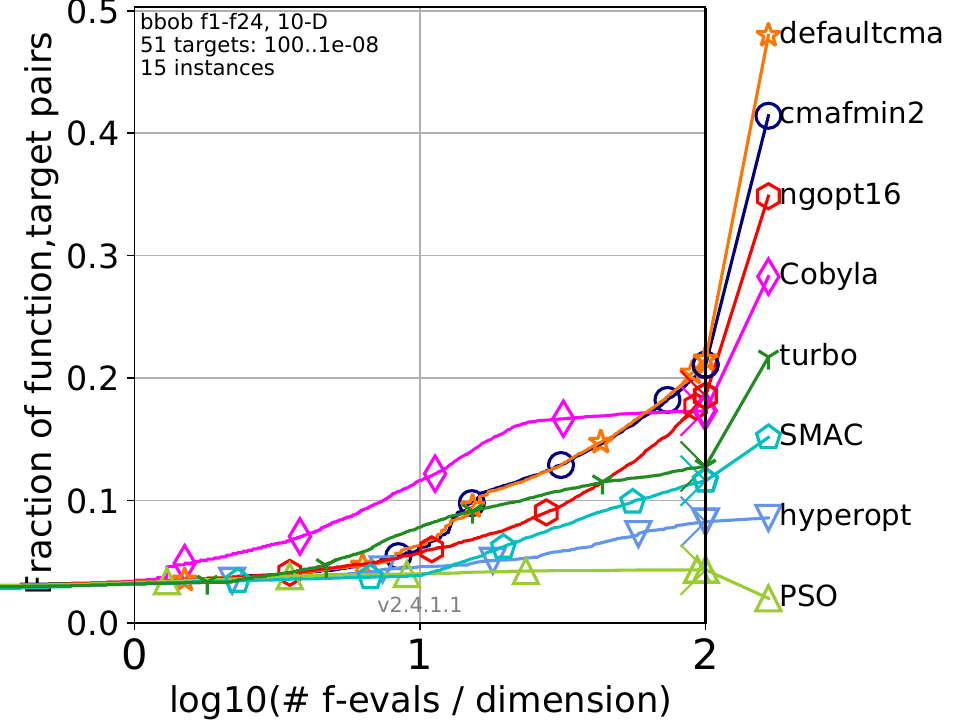}}
\caption{\label{b100d} As Fig. \ref{b10d} but with a $100D$ budget and only for dimensions (a) 2, (b) 3, (c) 5, and (d) 10. Some methods, which are too slow, are removed from the analysis. Overall, CMA algorithms and NGOpt16 perform best here.}
\end{figure*}

\textbf{Results for BBOB:} Figs.~\ref{b10d} and \ref{b100d} present results with budget equal to $10D$ and $100D$, respectively, in dimension~$D$.
The x-axis is the budget divided by the dimension, in logarithmic scale, while the y-axis shows the frequency of problems solved, i.e., the higher, the better. 
The plots are built by using the COCO/BBOB post-processing tool. It aggregates problems with different target precision values and displays the runtime distributions with simulated restarts~\cite{HansenABT22,hansen_coco_2021}. The default target precision values are 51 evenly log-spaced values between $10^{2}$ and $10^{-8}$. 
The complete data set is archived on Zenodo~\cite{zenodo} and allows further comparison with other BBOB data sets through the original COCO platform~\cite{hansen_coco_2021} or the IOHanalyzer web-interface~\cite{iohanalyzer}.

Our experiments for a budget of $10D$ reproduce the results in \cite{bbobsmac}, where SMAC outperforms CMA. 
However, SMAC performance decreases with increasing dimension. Similar to all BO-based algorithms, its computational complexity increases significantly with increasing dimension and with increasing budget.

Although we tested fewer optimizers for the $D= 20$ and $D=40$ cases due to the high computational cost, it is interesting to note that Cobyla, which is simply based on linear interpolation, often performs better than all other solvers.
Although there is no tuning for our present results from Cobyla on BBOB, it outperformed all BO-based algorithms. Cobyla is also the algorithm selected by the NGOpt wizard for the test cases considered, which explains their similar performance.

Fig. \ref{b100d} shows that both versions of CMA -- defaultcma and cmafmin2 (and NGOpt16, which uses CMA as a component) -- perform better than all other solvers when a larger budget of function evaluations is available. 

We conclude by remarking that, in the BBOB benchmark with our specific setting using a low budget, there are elements that have a significant impact on the overall result:
\begin{itemize}
    \item The LinearSlope function, which has optima in the corners, is difficult for methods that assume that the optimum is supposed to be inside. 
    \item The precision parameter, ranging from $1e-8$ to $100$ by default, has a significant impact on results. Since our budget is much smaller than the default setting, the percentage of the problems that are successfully solved is smaller than the typically reported ones. 
\end{itemize}
In an additional set of experiments, we verified that removing LinearSlope or changing the precision parameter does not change the overall picture of our results. 

\begin{figure*}[!h]
\centering
\subfloat[]{\includegraphics[angle=0,  clip, height=.17\textheight]{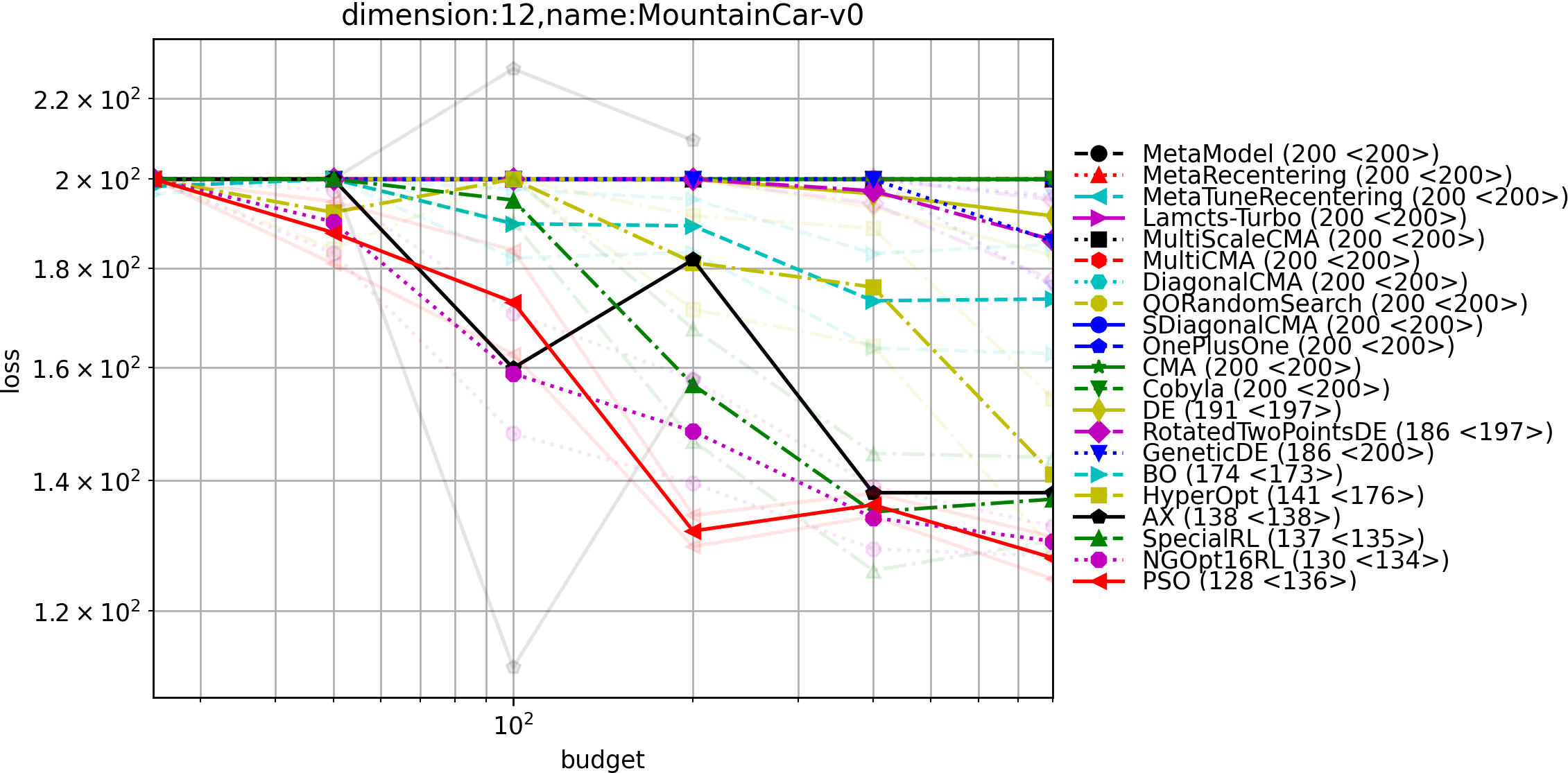}\label{fig:a}}\hfill
\subfloat[]{\includegraphics[angle=0,  clip, height=.17\textheight]{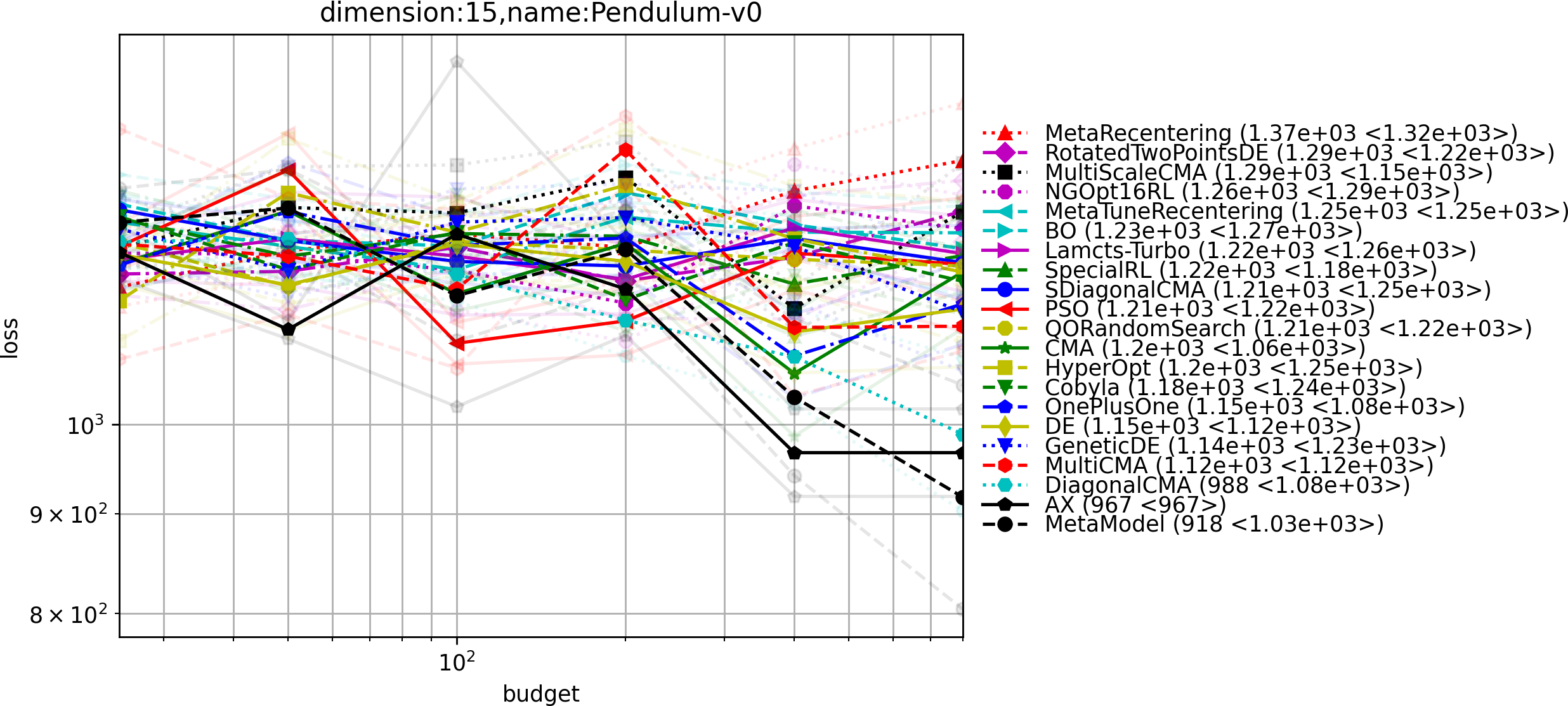}\label{fig:b}}\\
\subfloat[]{\includegraphics[angle=0,  clip, height=.17\textheight]{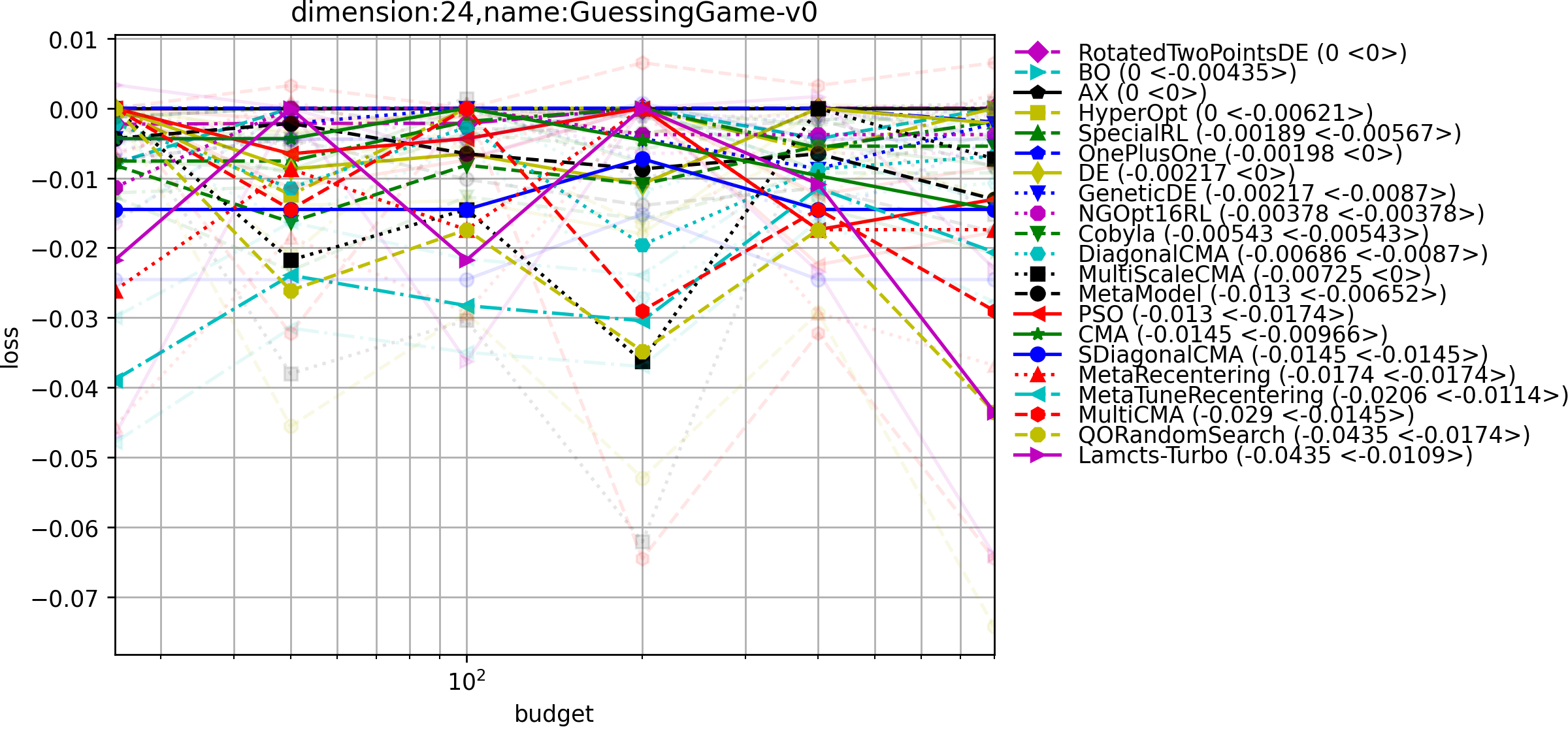}\label{fig:c}}\hfill
\subfloat[]{\includegraphics[angle=0,  clip, height=.17\textheight]{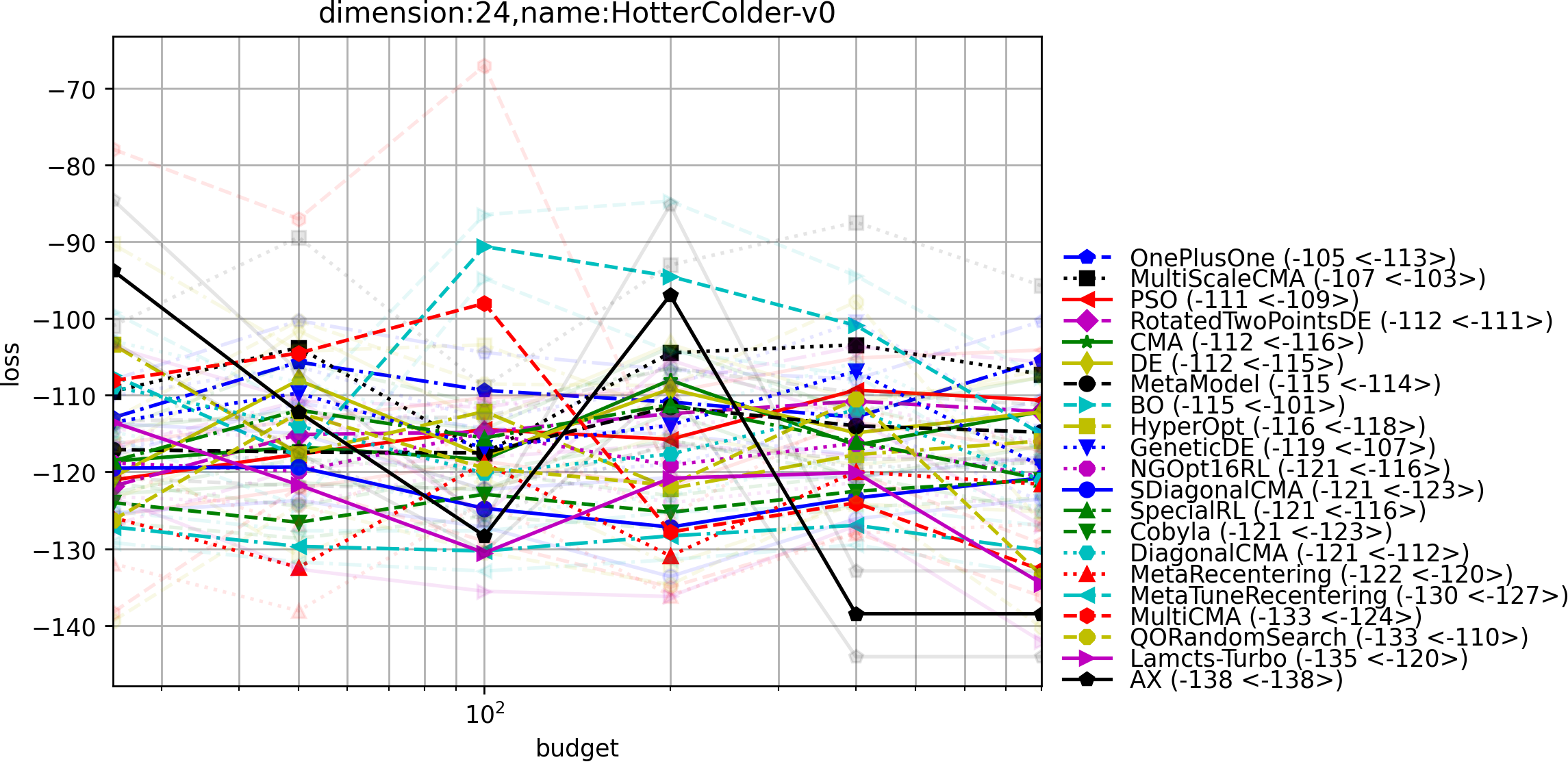}\label{fig:d}}\\
\subfloat[]{\includegraphics[angle=0,  clip, height=.17\textheight]{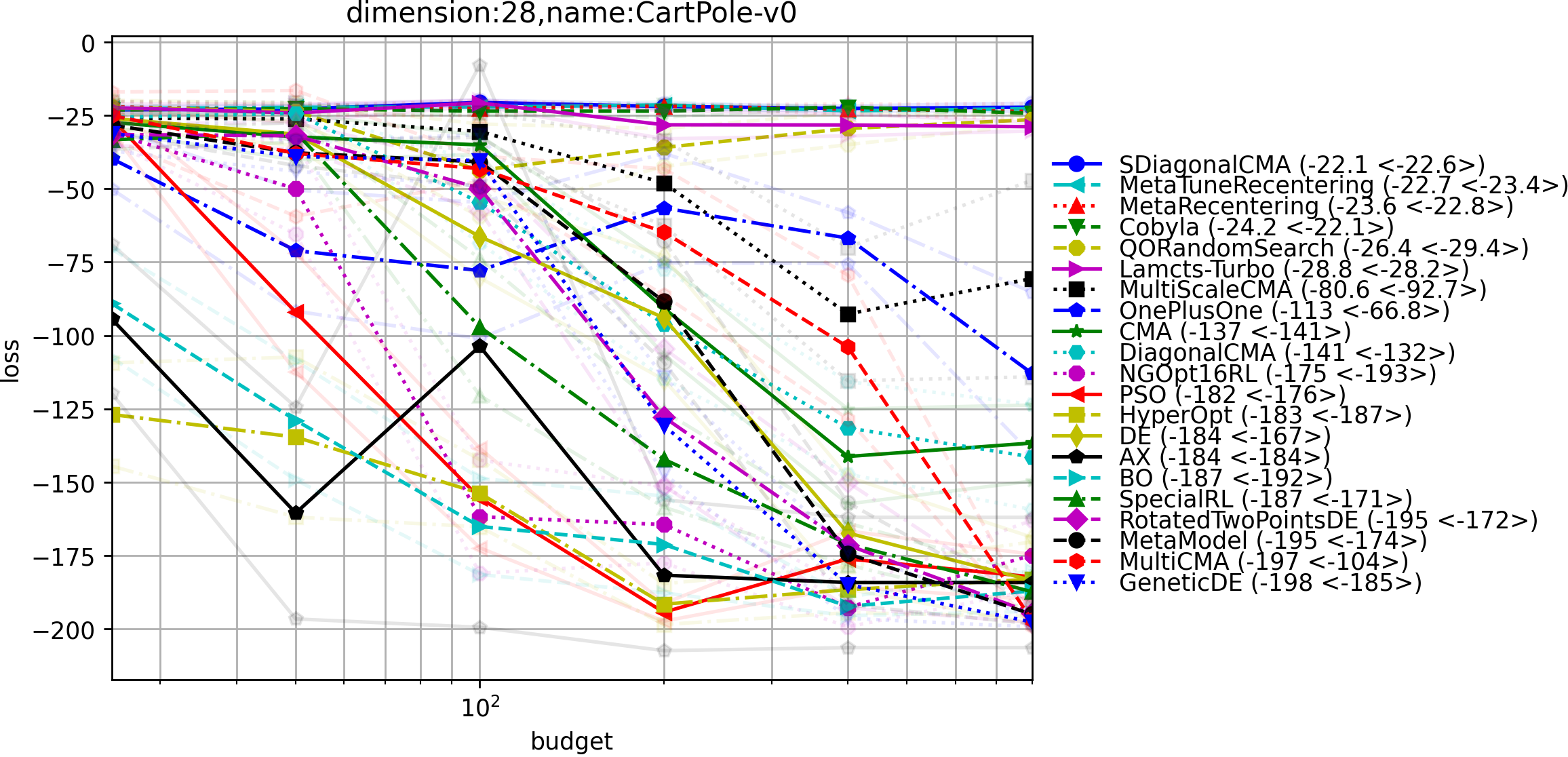}\label{fig:e}}\hfill
\subfloat[]{\includegraphics[angle=0,  clip, height=.17\textheight]{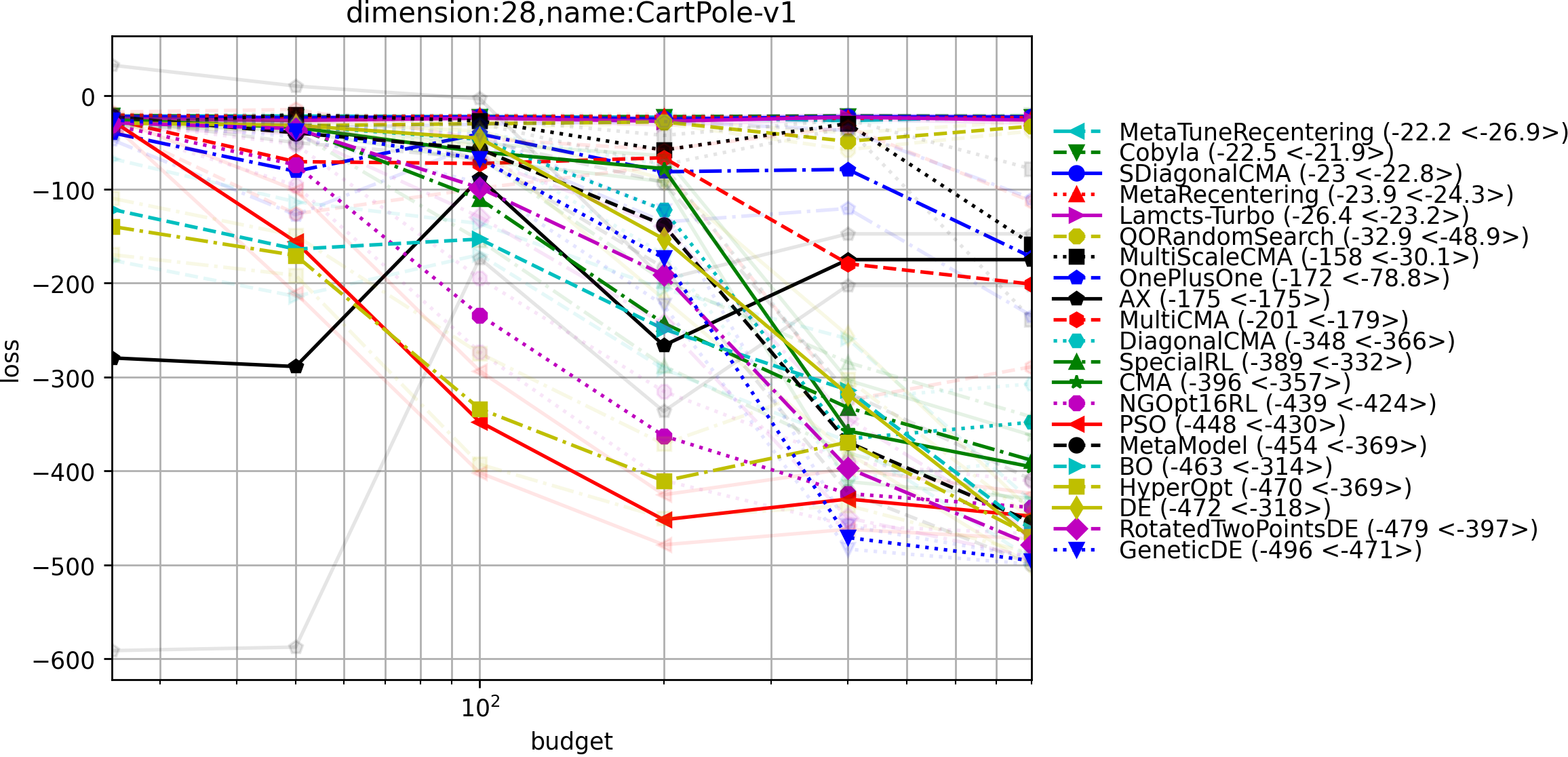}\label{fig:f}}\\
\subfloat[]{\includegraphics[angle=0,  clip, height=.17\textheight]{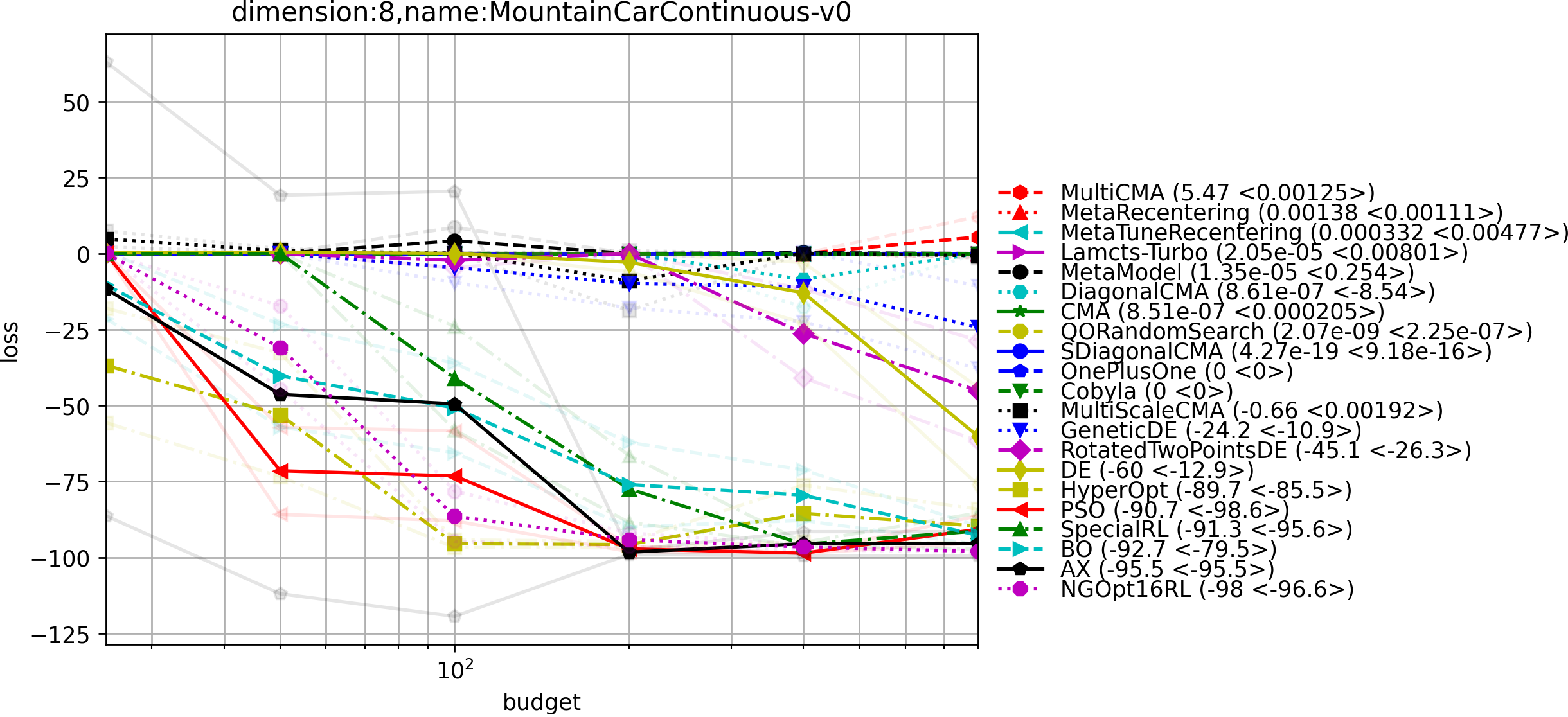}\label{fig:g}}
\subfloat[]{\includegraphics[angle=0,  clip, height=.17\textheight]{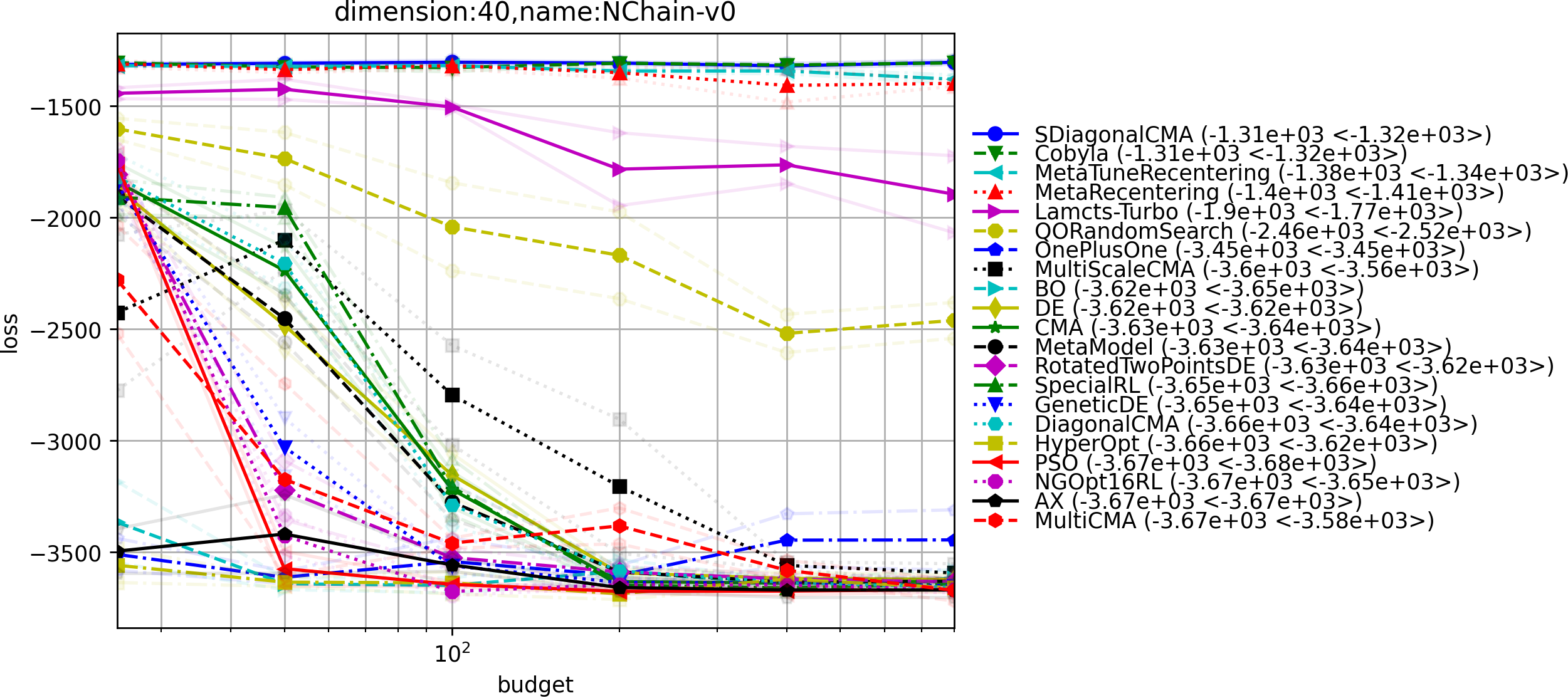}\label{fig:h}}\\
\caption{\label{det}Multi-deterministic Open AI Gym with tiny neural net: a random seed is randomly drawn for each optimization run so that overfitting is more difficult. See Fig. \ref{det2} for an aggregated view. The legend lists all the compared algorithms with two numbers in parentheses: These are the performance values for the last and second-to-last budgets on the x-axis, respectively. The first number is also used to sort the algorithms by performance.} 
\end{figure*}

Table~\ref{tab1} and Table~\ref{tab2} list the total execution time in seconds for 15 runs on the 24 BBOB functions for budget 10D and 100D, respectively. Note that NGOpt sometimes, in particular when the budget is sufficiently large compared to the dimension, spends a significant time learning a meta-model and checking in cross-validation whether this meta-model could be applied. NGOpt can hence be unexpectedly more expensive in lower dimensions, which in our experiments leads to the non-monotonic behavior of the running time with respect to the dimension.
DefaultCMA, Turbo20, and SMAC2 have comparable execution times to CMAFmin2, Turbo, and SMAC, respectively.
AX, BO, LAMCTS, and SMAC showed a total time of more than 20\,000 seconds in dimension 10. As their computational cost became unmanageable for dimension 20, we stopped the runs after 3-day wall-clock time. We therefore do not report results for these algorithms for the 20- and 40-dimensional BBOB functions with 10D budget. The same algorithms, with the exception of SMAC, plus Optuna, are excluded from the experiments with a higher total evaluation budget (100D) due to their excessive computational cost for that budget.

\begin{table}
\begin{center}
\caption{Total computational time in seconds for the BBOB benchmark Budget = 10D }
\label{tab1}
\begin{tabular}{| l | r | r | r | r | r | r |}
\hline
\textbf{Dimension} & \textbf{2} & \textbf{3} & \textbf{5} & \textbf{10} & \textbf{20} & \textbf{40} \\ 
\hline
CmaFmin2 & 30 & 31 & 33 & 38 & 50 & 90 \\ 
\hline 
AX & 32\,267 & 65\,044 & 63\,808 & 291\,845 &  &  \\ 
\hline
Turbo & 116 & 1\,118 & 3\,572 & 11\,535 & 58\,670 & 48\,700 \\ 
\hline
Cobyla & 16 & 24 & 32 & 51 & 86 & 202 \\
\hline
NGOpt & 90 & 204 & 303 & 563 & 167 & 336 \\ 
\hline
BO & 1087 & 2\,129 & 5\,486 & 23\,274 & &  \\ 
\hline
Optuna & 48 & 121 & 409 & 1\,733 & 7\,388 & 37\,431 \\ 
\hline
Lamcts & 389 & 1\,941 & 9\,804 & 47\,218 & &  \\ 
\hline
SMAC & 5\,978 & 9\,258 & 26\,295 & 66\,733 &  &  \\ 
\hline
HyperOpt & 34 & 117 & 420 & 1\,969 & 7\,716 & 21\,835 \\ 
\hline
PSO & 15 & 20 & 31 & 70 & 183 & 587 \\ 
\hline
\end{tabular}
\end{center}
\end{table}

\begin{table}
\begin{center}
\caption{Total computational time in seconds for the BBOB benchmark Budget = 100D}
\label{tab2}
\begin{tabular}{| l | r | r | r | r |}
\hline
\textbf{Dimension} & \textbf{2} & \textbf{3} & \textbf{5} & \textbf{10} \\ 
\hline
CmaFmin2 & 36 & 41 & 53 & 89 \\ 
\hline 
Turbo & 13\,241 & 22\,161 & 38\,903 & 141\,852 \\ 
\hline
Cobyla & 197 & 297 & 474 & 845 \\
\hline
NGOpt & 625 & 1\,254 & 3\,070 & 333 \\ 
\hline
SMAC & 224\,031 & 205\,209 & 351\,508 & 807\,572 \\ 
\hline
HyperOpt & 1\,038 & 1\,655 & 4\,260 & 17\,836 \\ 
\hline
PSO & 60 & 101 & 179 & 458 \\ 
\hline
\end{tabular}
\end{center}
\end{table}

\textbf{Results for the Direct Policy Search on OpenAI Gym:} In Fig.~\ref{det}, we provide an independent comparison of black-box solvers applied to the Ng-Full-Gym benchmark, which is Nevergrad's direct policy search applied to OpenAI Gym. More precisely, is the optimization of a neural network as a controller for OpenAI Gym problems. We plot the unscaled loss, defined as the opposite of the reward the agent accumulates over time. Here, the lower the curve, the better the performance.
The non-monotonic trend of the curves with respect to the elapsed budget is due to the fact that, for a given problem and algorithm, we perform new runs each time the total budget of the evaluations is updated. As a consequence, the algorithm may choose a different parametrization depending on the total budget or the ratio between the budget and the dimension of the specific benchmark, leading to a statistically significant difference in performance. In addition, since we perform independent runs for different budgets and average over the repetitions, some variability in performance is to be expected.

We add the following algorithms to our comparison:
QORandomSearch \cite{quasiopposite}, MetaTuneRecentering \cite{ppsnrescaling}, and MetaRecentering \cite{CauwetCDLRRTTU20} are variants of random search that are fully parallel and reduce redundancies compared to random search. 
We also add NGOpt16RL and SpecialRL from Nevergrad: NGOpt16RL is Nevergrad's wizard NGOpt optimized specifically for RL problems; SpecialRL runs the base NGOpt16RL for half of the budget and then uses test-based population size adaptation (TBPSA)~\cite{vasilfoga} in the second half.
Moreover, since the OpenAI Gym suite contains problems that are unbounded with optima at a small scale, we add algorithms that can adjust the scale of the search. GeneticDE and RotatedTwoPointsDE, for example, are designed to extrapolate the scale from some variables to others and find the right scale before running the classic DE~\cite{de} algorithm. MetaTuneRecentering performs a DoE entirely designed to choose the right scale depending on the relationship between budget and dimension. MultiCMA is a tentative robustification of CMA that runs three copies of CMA.
 We also consider a (1+1) evolutionary algorithm~\cite{oneplusone}, which is a simple and fast state-of-the-art heuristic for adjusting the scale, Diagonal CMA~\cite{cma}, which uses a diagonal covariance matrix, and its scaled version, Scaled Diagonal CMA~\cite{nevergrad}.
Finally, we add to our portfolio Lamcts-Turbo~\cite{lamctsNeurips}, an improved version of vanilla LA-MCTS coupled with Turbo to draw new samples in the selected node of the Monte Carlo search tree at each iteration. For a complete list of all algorithms compared, see Section VI in the Supplementary Material.

\begin{figure*}[b]
\centering
\vspace{-0.0cm}\subfloat[Budget 25]{\includegraphics[angle=0,  clip,width=.45\textwidth]{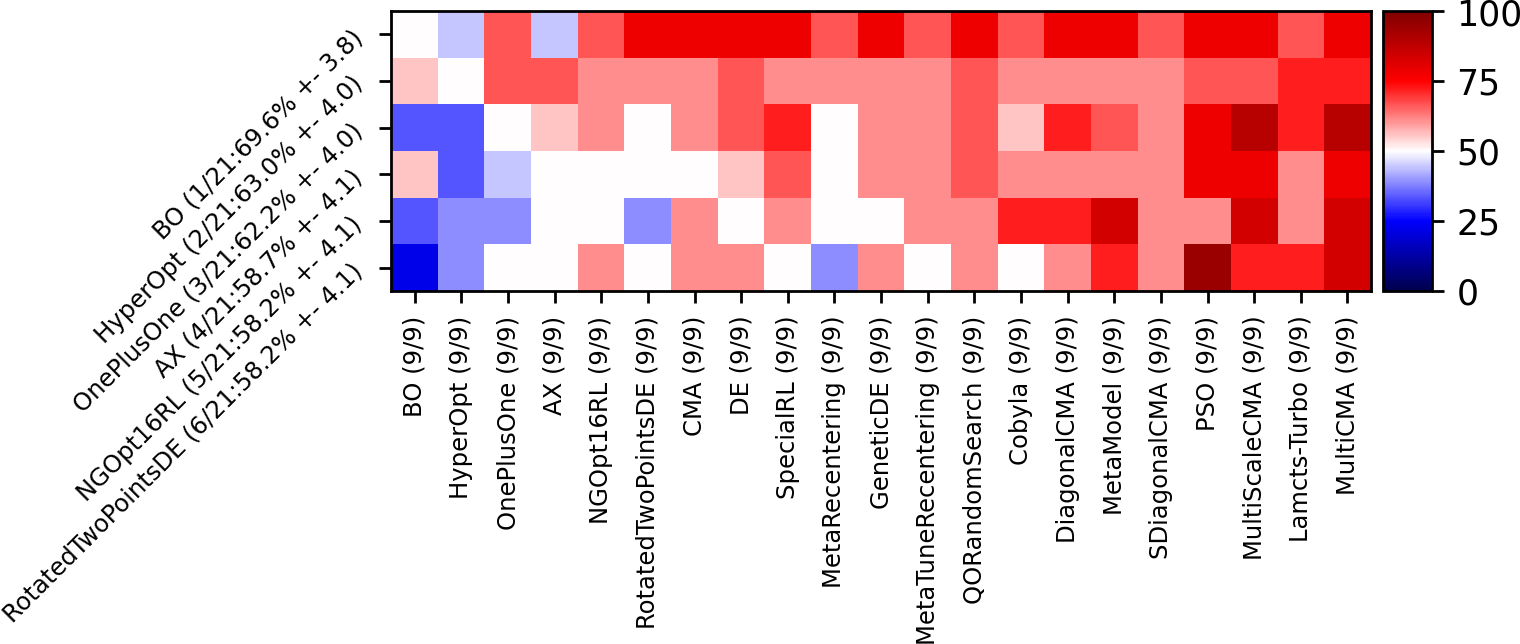}}
\vspace{-0.0cm}\subfloat[Budget 50]{\includegraphics[angle=0,  clip,width=.45\textwidth]{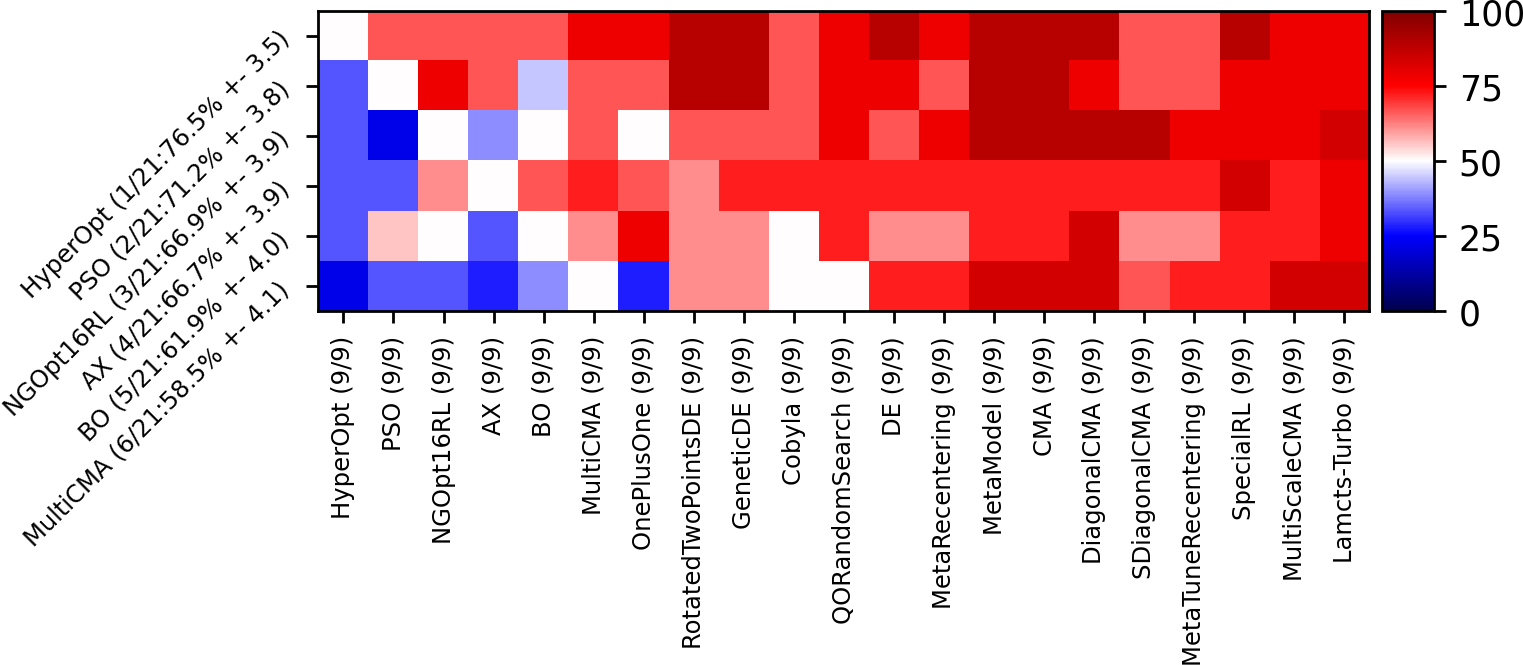}}\\
\vspace{-0.0cm}\subfloat[Budget 100]{\includegraphics[angle=0,  clip,width=.45\textwidth]{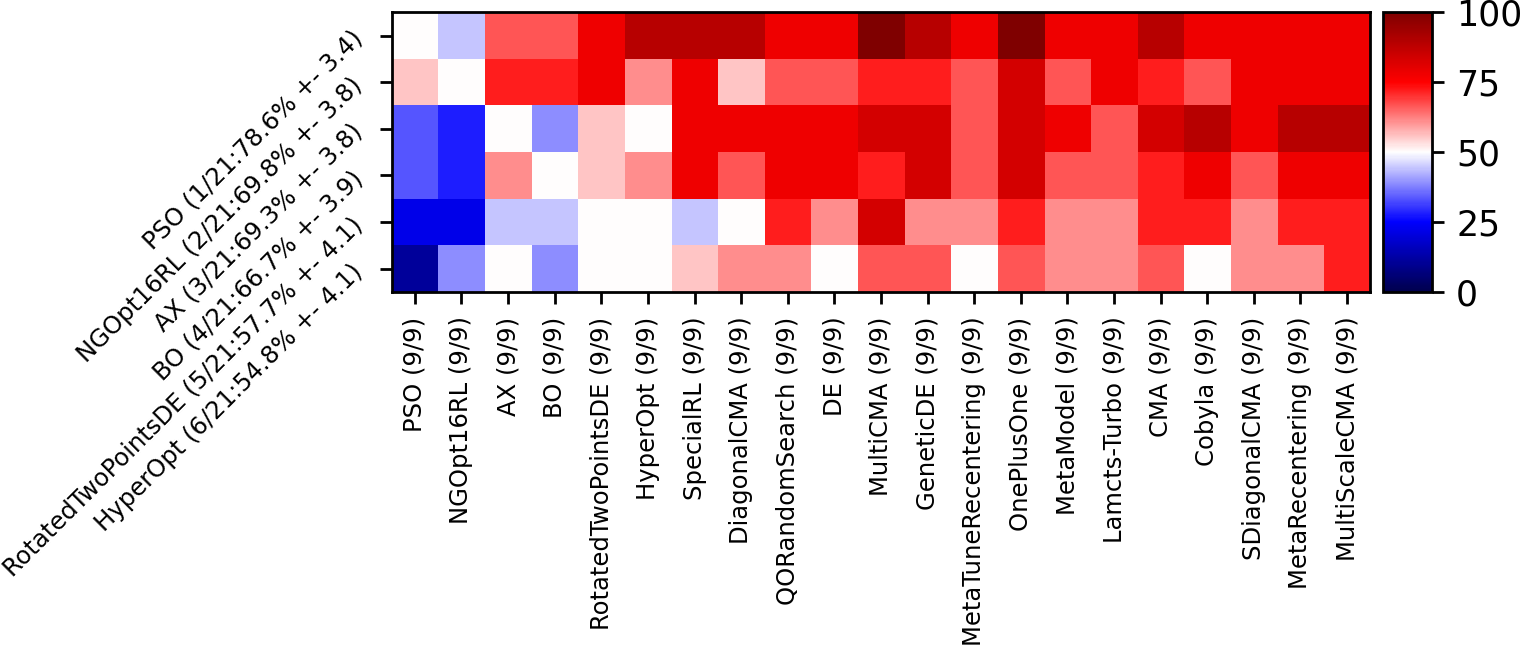}}
\vspace{-0.0cm}
\subfloat[Budget 200]{\includegraphics[angle=0,  clip,width=.45\textwidth]{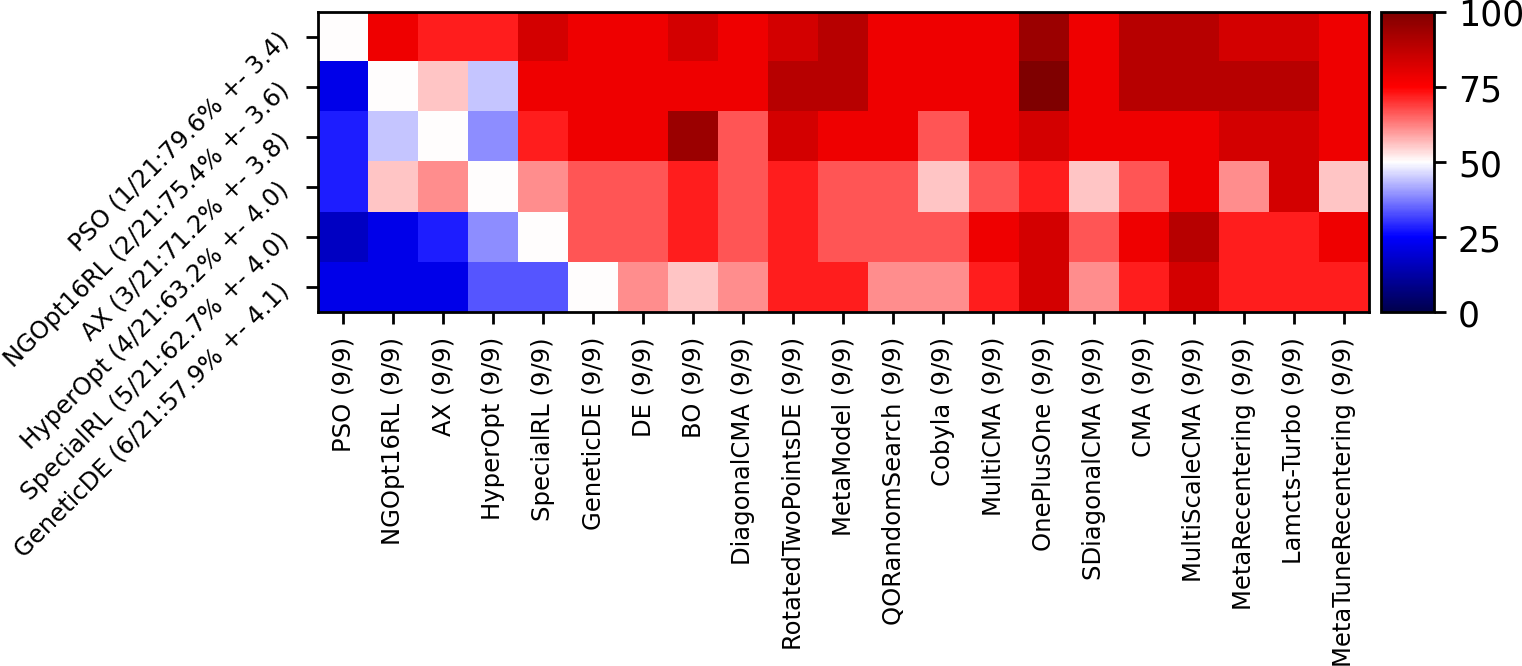}}\\
\vspace{-0.0cm}\subfloat[Budget 400]{\includegraphics[angle=0,  clip,width=.45\textwidth]{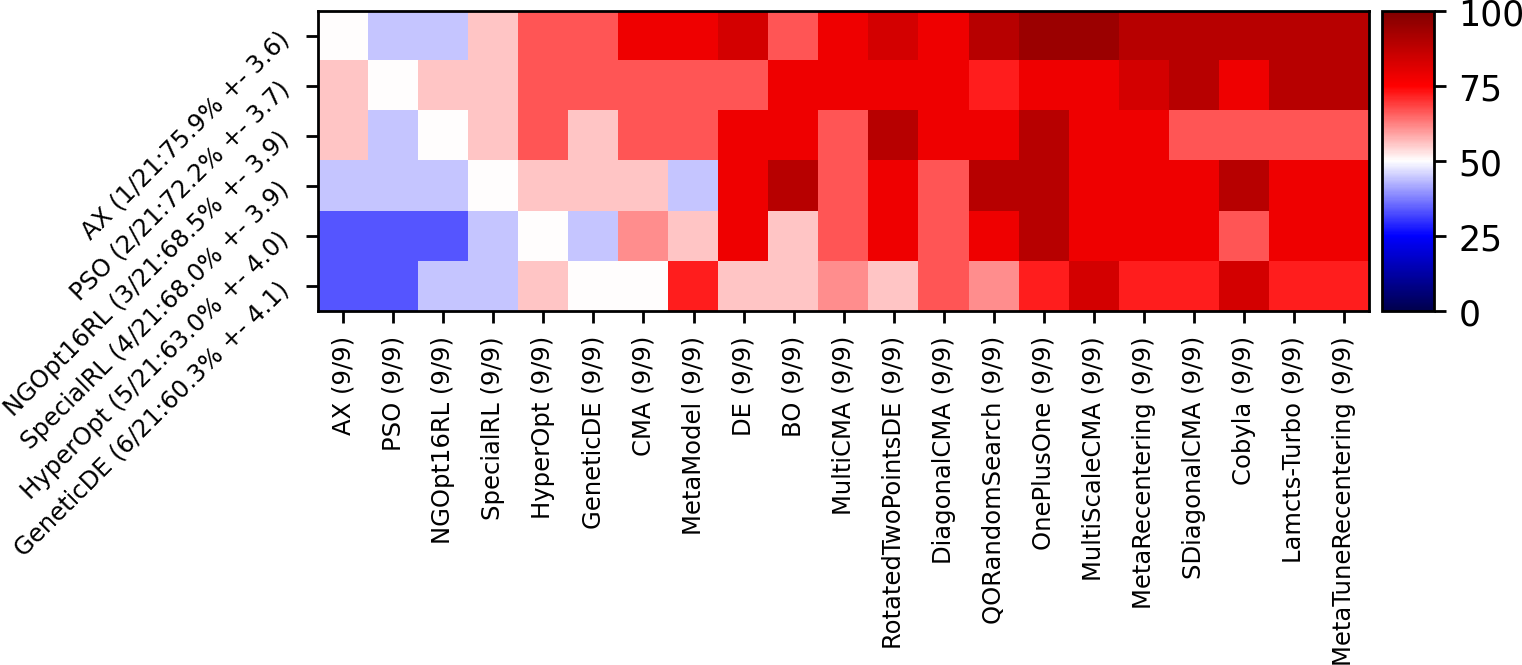}}
\vspace{-0.0cm}\subfloat[All budgets]{\includegraphics[angle=0,  clip,width=.45\textwidth]{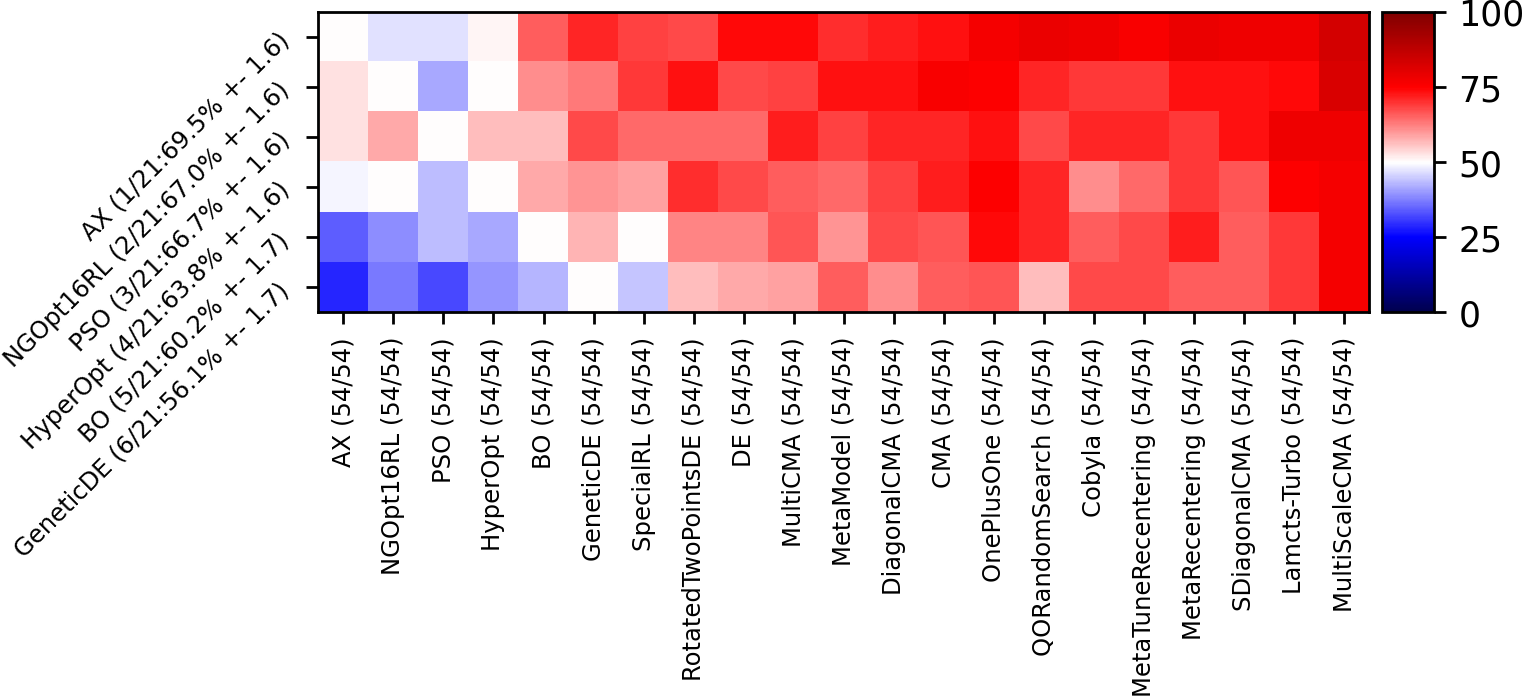}\label{fig:all_budgets}}
\\
\caption{\label{det2}Same results as reported in Fig. \ref{det}, but aggregated comparison as provided by Nevergrad~\cite{nevergrad}. Row A column B shows the frequency at which method A outperformed method B for the given budget. 9 distinct problems per budget. We include only problems for which the dimension is $D<50$. Methods are ranked per average winning rate; best methods are listed first. Note that winning rates are all very close to each other: only PSO is significantly better. Fig.~\ref{det3} presents similar experiments but with bigger neural nets. Fig. 1 in the Supplementary Material extends the present results to budgets 800, 1600, and 3200.}
\end{figure*}

Although we cannot derive comprehensive recommendations from Fig.~\ref{det} because of the high variability of results across different benchmarks and dimensions, some conclusions can be drawn from the observation of similar patterns.
First, we note that Fig. \ref{fig:b}, \ref{fig:c}, and \ref{fig:d} corresponding to problems of intermediate dimension $(D = 15$ and $D = 24)$ show no significant differences in solver performance.
On the contrary, for the low-dimensional test cases in Fig. \ref{fig:a} and \ref{fig:g}, we observe a very clear superiority of the BO-based solvers (BO and AX), HyperOpt, PSO, and the wizard NgOpt16RL, all consistently belonging to the 5-best group. On the other hand, Fig. \ref{fig:e}, \ref{fig:f}, and \ref{fig:h} show that the quality of the performance of the 5-best group deteriorates as the dimension of the problem increases. 
Indeed, we observe the lowest loss values at the end of the total evaluation budget for population-based algorithms like DE and CMA, which are known to be powerful heuristics to address difficult black-box problems when a large number of function evaluations are available.
However, for lower budgets of up to 100 function evaluations, the best-performing algorithms are still BO, AX, HyperOpt, PSO, and NgOpt16RL. On the contrary, Lamcts-Turbo consistently performs poorly on OpenAI Gym benchmarks, regardless of problem budget and dimension, in sharp contrast to the results shown in~\cite{lamctsNeurips}, where LA-MCTS was reported to perform well on some OpenAI Gym problems. We suspect that this discrepancy is caused by a poor initialization of the competitors in~\cite{lamctsNeurips}.

\begin{figure*}[t]\centering
\vspace{-0.0cm}\subfloat[Budget 25]{\includegraphics[angle=0,  clip,width=.42\textwidth]{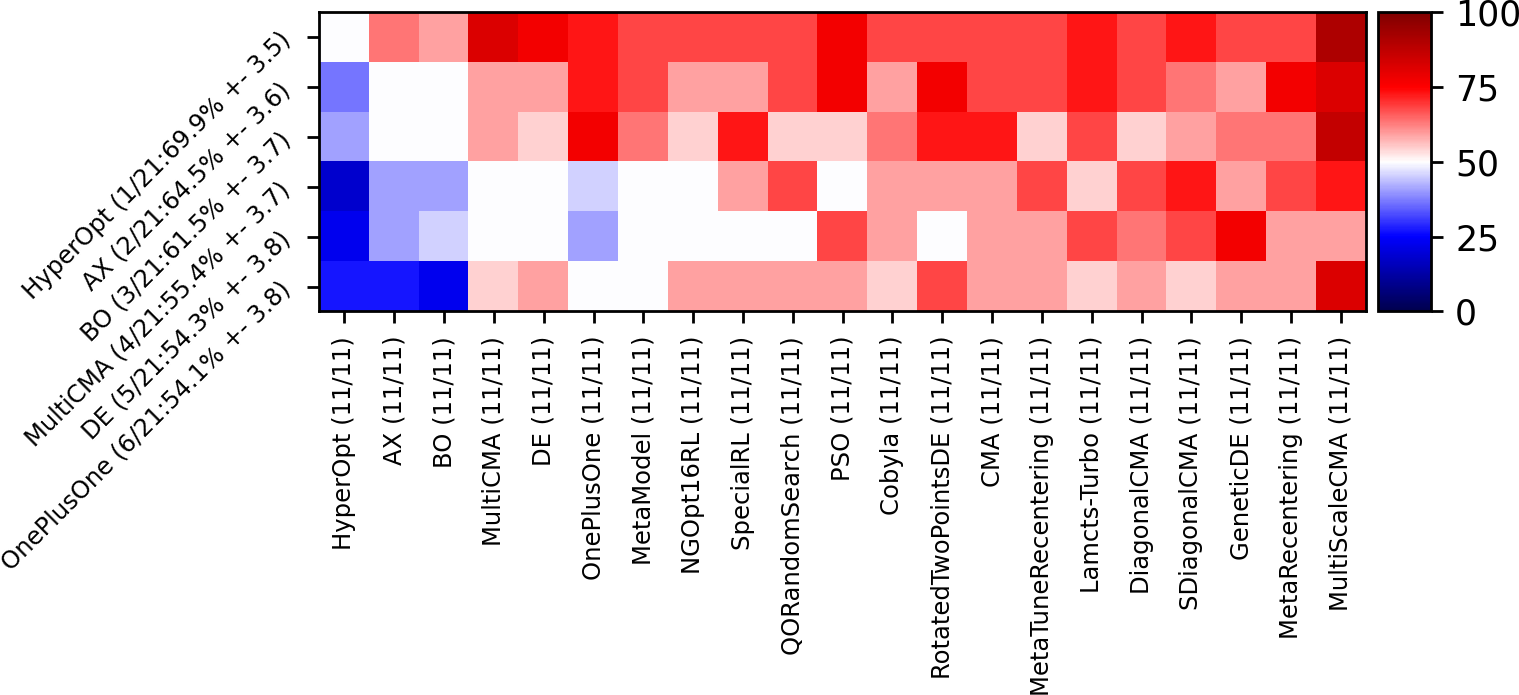}}
\vspace{-0.0cm}\subfloat[Budget 50]{\includegraphics[angle=0,  clip,width=.42\textwidth]{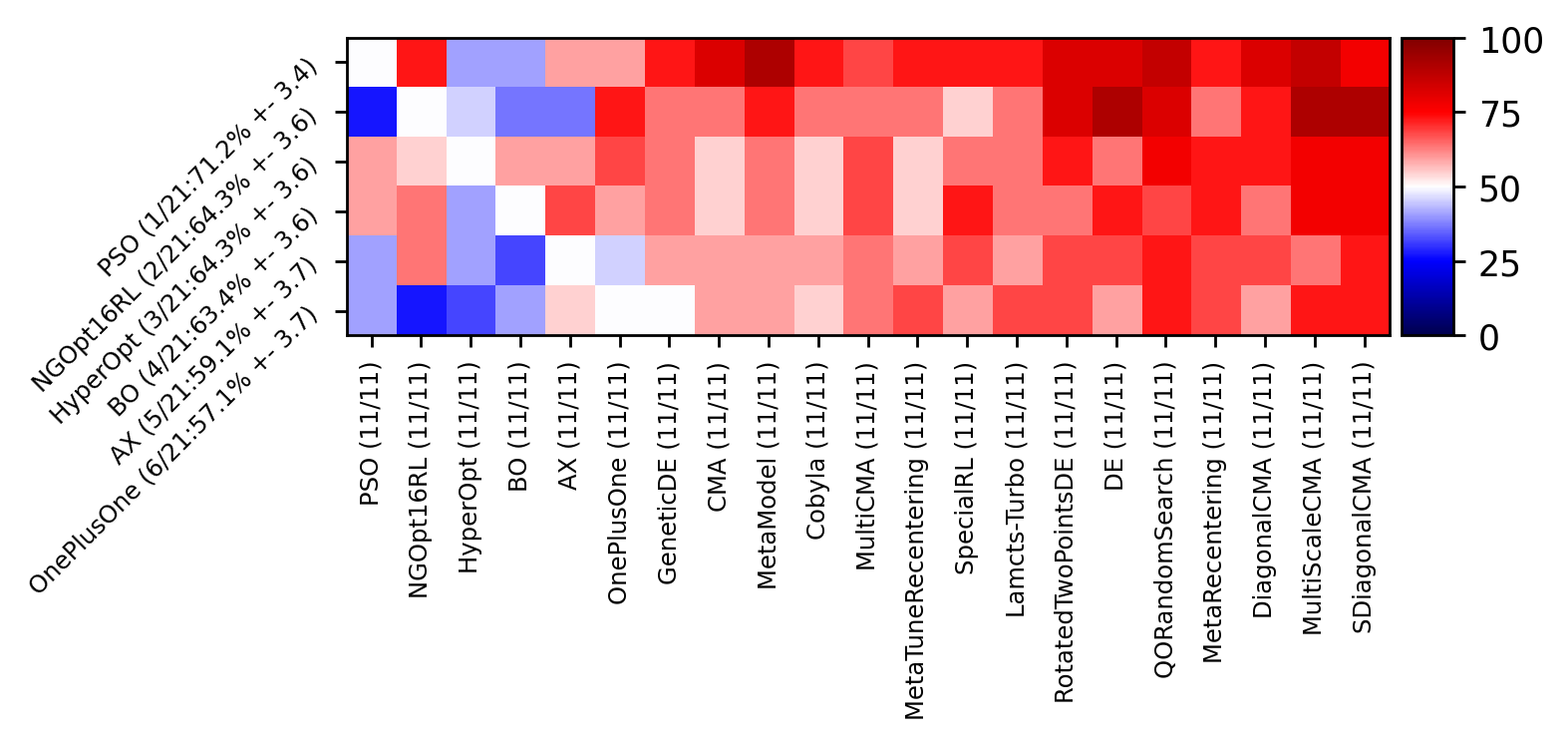}}\\	\vspace{-0.0cm}\subfloat[Budget 100]{\includegraphics[angle=0,  clip,width=.42\textwidth]{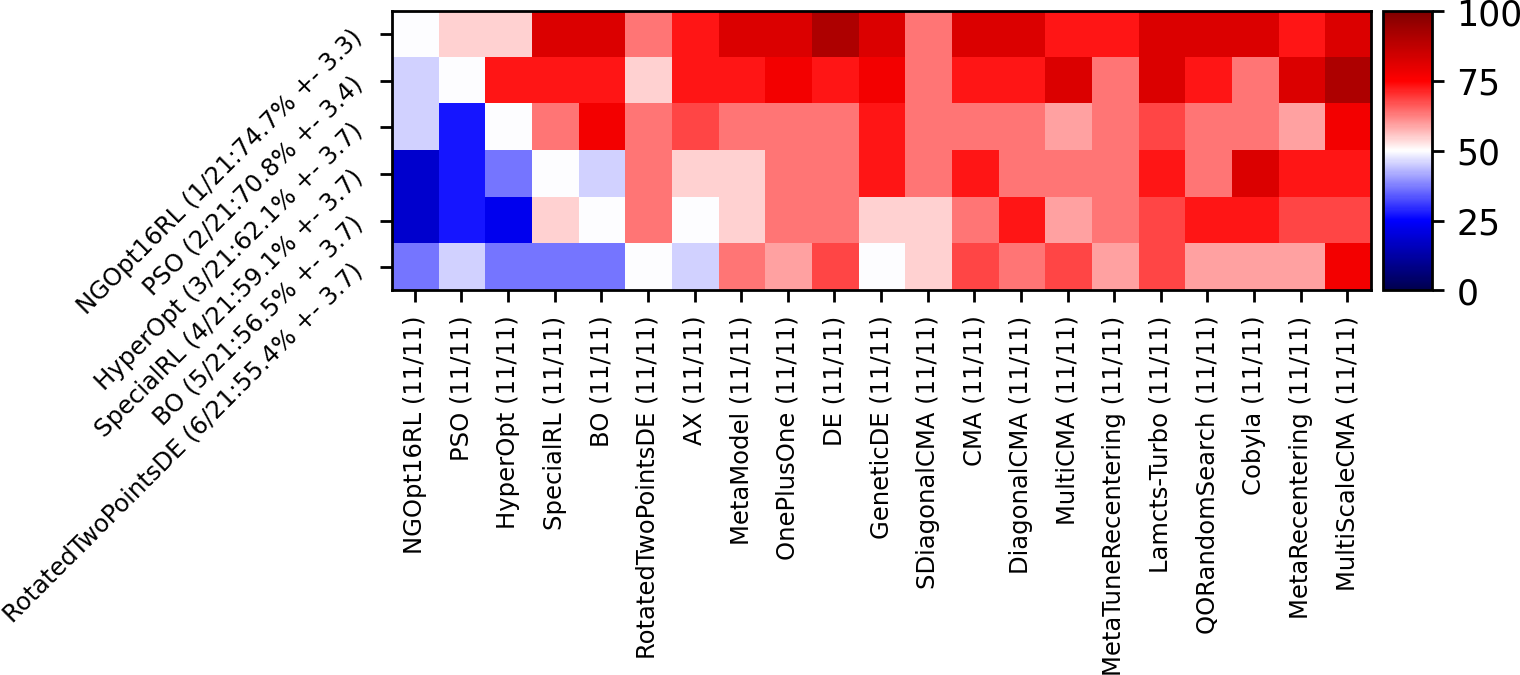}}
\vspace{-0.0cm}\subfloat[Budget 200]{\includegraphics[angle=0,  clip,width=.42\textwidth]{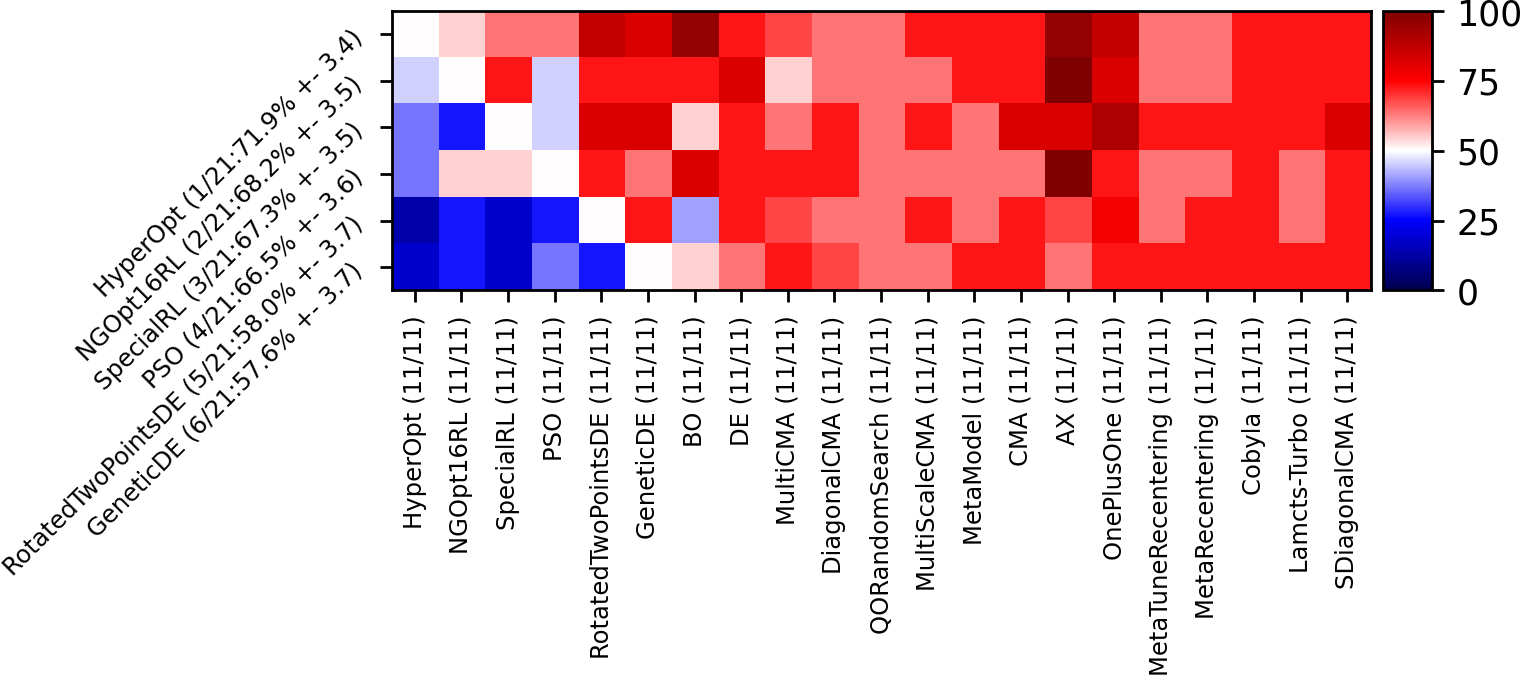}}\\
\vspace{-0.0cm}\subfloat[Budget 400]{\includegraphics[angle=0,  clip,width=.42\textwidth]{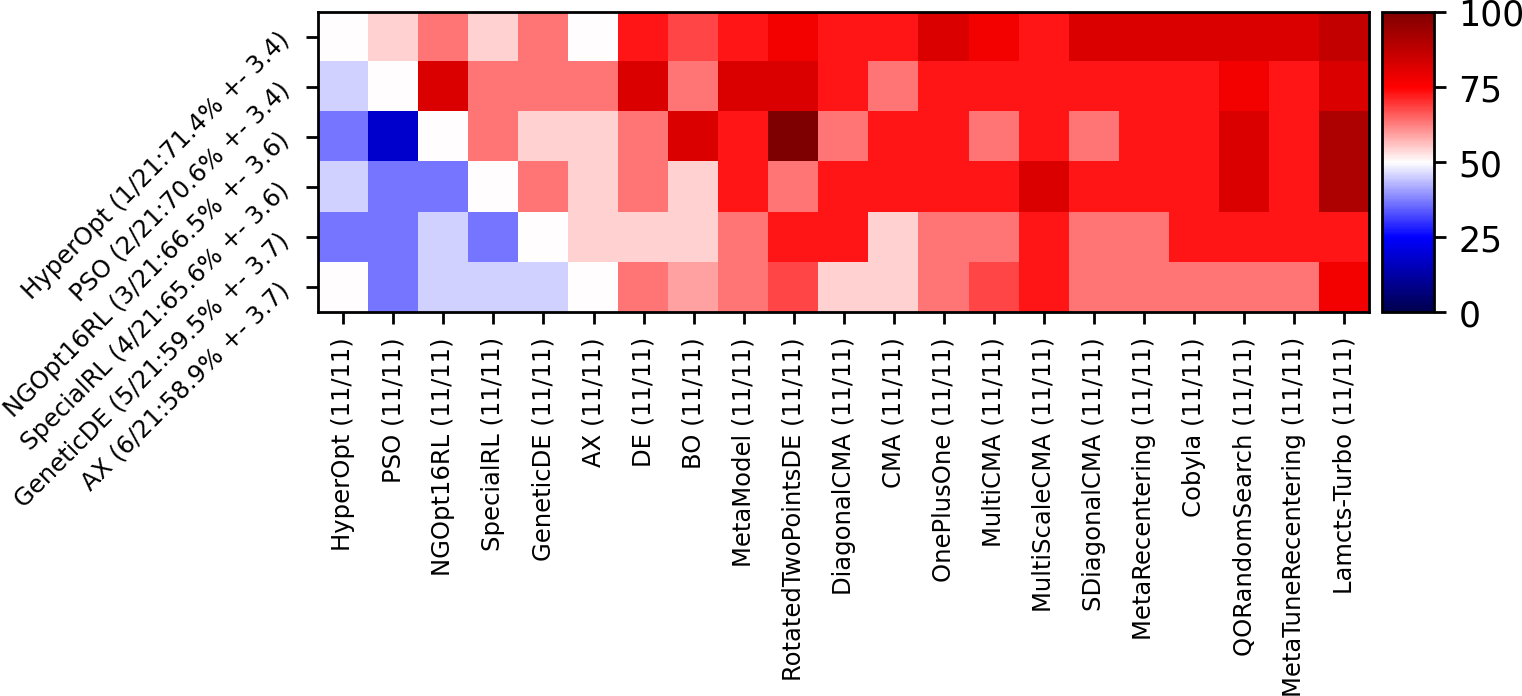}}
\vspace{-0.0cm}\subfloat[All budgets]{\includegraphics[angle=0,  clip,width=.42\textwidth]{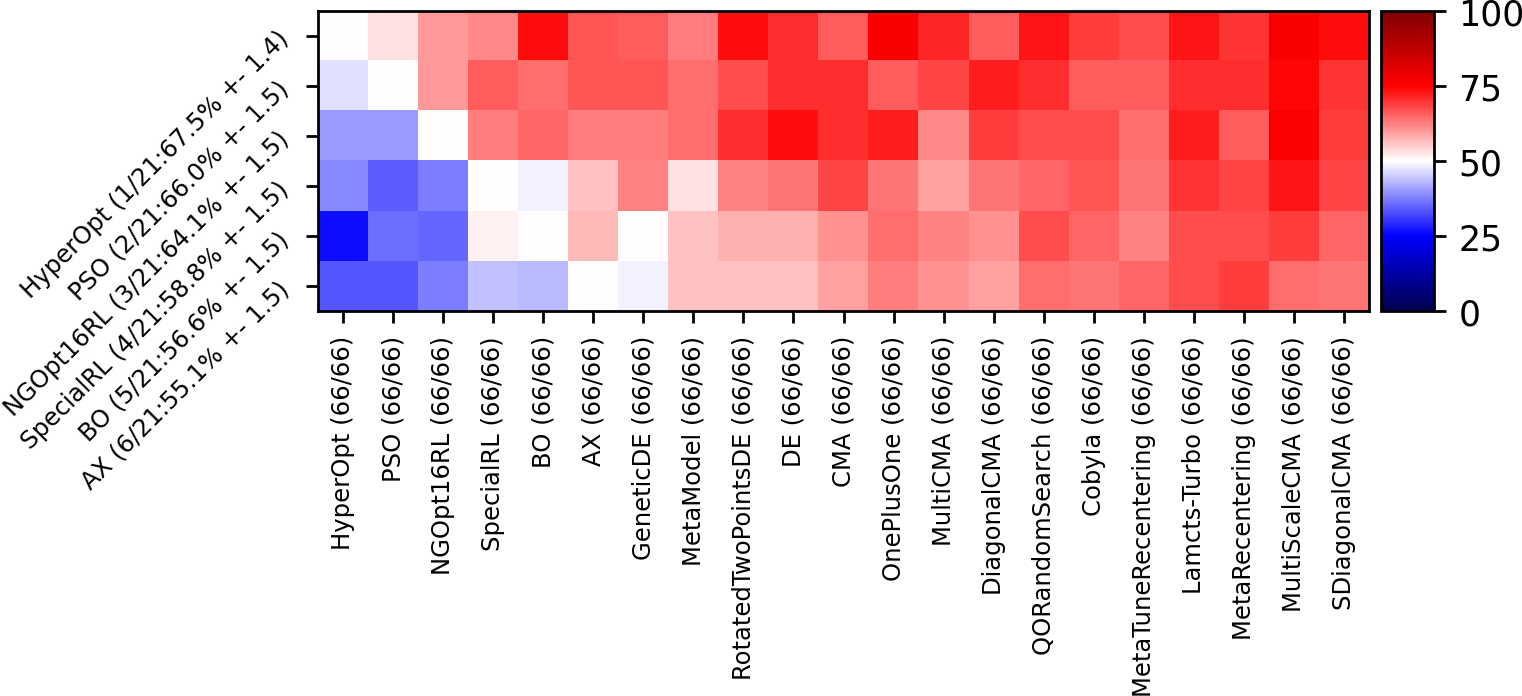}\label{fig:big_all_budgets}}\\
\caption{\label{det3}Same as Fig. \ref{det2}, but with bigger nets (neural factor 3 in Nevergrad's benchmark scaling). 11 distinct problems per budget. We truncated at dimension $\leq 264$. Dimension ranges from 24 to 264 instead of 8 to 40 in Fig. \ref{det2}. Due to the computational cost, it was not possible to finish the runs for SMAC. Fig. 1 in the Supplementary Material extends the present results to budgets 800, 1600, and 3200.} 
\end{figure*}

For an in-depth comparison, we also present aggregate plots based on average winning rates (e.g., Fig. \ref{det2}), where we observe good results for the BO-based methods: BO is the most powerful solver for the lowest budget, while AX, NGOpt16RL and PSO turn out to be the best algorithms (within the confidence intervals) when aggregating results across all budgets (Fig. \ref{fig:all_budgets}). This means that vanilla BO is actually good for fast low-precision approximations on difficult problems, whereas AX, NGOpt16RL and PSO can be recommended when optimizing low-dimensional problems with low to medium budgets. Furthermore, Fig. \ref{det2} confirms, on the one hand, the good performance of NGOpt16RL, PSO, and HyperOpt for benchmarks with tiny neural networks, and, on the other hand, the poor search capabilities of Lamcts-Turbo, regardless of the available budget.
For higher dimensions, Fig.~\ref{det3} shows that BO-based solvers become less competitive, although they still perform among the best for the smallest budgets (25 and 50 evaluations). For larger budgets, we see that CMA and DE start becoming more competitive, which is in line with \cite{MeunierRWRRTMD22}. However, Fig.~\ref{fig:big_all_budgets} shows a supremacy of HyperOpt, which is designed for large-scale optimization for models with hundreds of parameters, and PSO, which performs surprisingly well considering that it is a classical heuristic originating from the BBO field and rarely used in ML applications.
It can also be noted that Cobyla does not perform as well as for BBOB, while PSO performs constantly well for intermediate budgets from 50 to 400 total evaluations. 

Fig. 1 in the Supplementary Material extends the present results to budgets 800, 1600, and 3200 for both tiny and big neural nets. Here, as the total budget increases, a smaller set of algorithms are compared: AX is affordable for runs on tiny neural nets up to a budget of 800 evaluations, and BO and Lamcts-Turbo become too expensive when exceeding a total budget of 1600 evaluations for big neural nets.
The RL-specific algorithms, SpecialRL, and NGOpt16RL achieve the best performance for the largest budgets, followed by the various CMA and DE versions, leaving PSO and HyperOpt behind.

All in all, our results on OpenAI Gym show the competitiveness of BO-based methods for small dimensions and evaluation budgets (up to 100 evaluations), comparably to other solvers, while solvers from other families perform better as the budget increases. We note that this is different behavior from the BBOB benchmarks, where BO-based methods never rank first and show only average performance.

\section{Conclusion}
Our results provide insight into the comparison between different state-of-the-art BO methods, commonly used in ML, and more classical BBO heuristics. We compared the solvers on the BBOB benchmark suite from the COCO environment, which is well-known in the BBO community, and on OpenAI Gym problems, which are from the RL domain.
On the BBOB comparison, we noted that Cobyla and the Nevergrad wizard perform best for any tested dimensionality from 2D to 40D with a total budget of 10D. They were consistently better than SMAC and Turbo, which was highlighted as a strong BO method in~\cite{TurnerEMKLXG20}. It is worth noting that Turbo and HyperOpt have the advantage of being computationally cheaper than other BO methods. Moreover, HyperOpt performed among the best on OpenAI Gym, with the exception of the largest considered budget for big neural nets. Good performance was also observed for PSO, which is commonly considered a classical heuristic and hardly ever used for ML tasks.
For the larger budget of 100D, CMA or the Nevergrad wizard perform best regardless of the considered dimension of the problem. 

In general, we found that the OpenAI Gym benchmark is very sensitive to variable scaling. While BBOB focuses on translations of optima and might therefore favor algorithms tuned for this setting, direct policy search for OpenAI Gym has an unbounded domain and depends differently on the initialization scaling and the ability of the algorithm to change scaling as needed. We optimized the scaling to Bayesian Optimization methods for obtaining results in Fig. \ref{det2}, and then we ran the experiments to get the (possibly more neutral) results in Fig. \ref{det3} without any change. 

Overall, tools based on heavy use of machine learning are computationally more expensive and perform roughly equivalently to mathematical programming or evolutionary techniques. In future work, we plan to compare BO and other methods in discrete settings as well. Our results also indicate a high relevance of initializing the solvers with the right scaling. Identifying suitable methods for a dynamic control policy is therefore another topic that we aim to investigate in future works. Of course, it remains interesting to periodically update our comparisons with new state-of-the-art solvers. Since all our experiments are performed with Nevergrad, such an ongoing benchmarking is largely facilitated: users can simply add their favorite method and compare their results to the ones reported above. 

\section*{Acknowledgments}
Our work is supported by ANR, project ANR-22-ERCS-0003-01, by the CNRS INS2I emergence project RandSearch, and by the PRIME programme of the German Academic Exchange Service (DAAD) with funds from the German Federal Ministry of Education and Research (BMBF).



\appendix

\section{Discussion: differences between BBOB and Nevergrad's YABBOB}\label{diffs}
There are differences between the benchmarking suites.

\subsection{Domains}
First, working in unbounded domains with a Gaussian distributed optimum (as in some benchmarks in Nevergrad) leads to differences compared to benchmarks on bounded domains such as BBOB, especially for functions with an optimum at the boundary. 
Both contexts look interesting, but some adaptation of NGOpt was needed to make it tackle optima on the boundary. In addition, we add bounded counterparts of YABBOB termed YABOUNDEDBBOB and YABOXBBOB for facilitating further research with simultaneously the convenience of Nevergrad (see Section \ref{para}) and the bounded setting of BBOB.

\subsection{Budgets and independence}
In BBOB, when specifying a maximum budget of $100D$, the results plotted for a lower budget of $10D$ are obtained as a truncation of the $100D$-budget runs.
On the contrary, Nevergrad runs each budget separately, which is more expensive from a computational point of view but gives a more complete picture of the different budgets. This can be remedied by launching distinct runs for different budgets for BBOB.

\subsection{Parallelization}\label{para} 
Nevergrad was more suitable than BBOB for testing very slow algorithms like AX \cite{ax} because it is easy to massively parallelize it on a cluster.

\section{BBOB interfaces}\label{interfacing}
A strength of BBOB is that the interfacing is quite easy, greatly facilitating reproducibility~\cite{reproducibilityTELO}.
\begin{lstlisting}[language=Python]
% Nevergrad's NGOpt.
% Also SMAC, SMAC2, AX, BO: using Nevergrad's API.
ng.optimizers.NGOpt(ng.p.Array(lower=lbounds, upper=ubounds, shape=[dim]), num_workers=1, budget=evals).minimize(f)

% HyperOpt.
fmin(fn=lambda x: f([x['w'+str(i)] for i in range(dim)]), space={'w'+str(i): hp.uniform('w' + str(i), -5, 5) for i in range(dim)}, algo=tpe.suggest, max_evals=evals)

% Optuna.
class OptunaObjective(object):
    def __init__(self, problem):
        self.problem = problem

    def __call__(self, trial):
        x = []
        for i in range(self.problem.dimension):
            x.append(trial.suggest_float("x{}".format(i), problem.lower_bounds[i], problem.upper_bounds[i]))
        return self.problem(x)

study = optuna.create_study(direction="minimize")
study.optimize(OptunaObjective(problem), n_trials=evalsleft())

% Turbo.
class turbo_function:
    def __init__(self, dim=len(lbounds)):
        self.dim = dim
        self.lb = lbounds
        self.ub = ubounds

    def __call__(self, x):
        assert len(x) == self.dim
        assert x.ndim == 1
        assert np.all(x <= self.ub) and np.all(x >= self.lb)
        return f(x)

my_turbo = Turbo1(f=turbo_function(), lb=lbounds, ub=ubounds, max_evals=evals, n_init=min(evals, 20))
    
my_turbo.optimize()

% LA-MCTS.
% We tested several successive variants of the code,
% without much impact: the version below is the last. 
% We also tested several values
% of ninits, without much change.

agent = MCTS(lb = f.lb,     
            ub = f.ub,
            dims = f.dims,
            ninits = 40,  # We tested variants without much change.
            func = f
            )
agent.search(iterations = evalsleft())
\end{lstlisting}

\section{How to run the experiments on OpenAI Gym}
\label{app:openAIgym_interface}
OpenAI Gym was recently introduced in Nevergrad. After cloning Nevergrad at
\url{https://github.com/facebookresearch/nevergrad.git}, experiments can be launched with the following command line: 

\begin{lstlisting}
python -m nevergrad.benchmark ng_full_gym --repetitions=10 --plot
\end{lstlisting}

However, in order to run a reduced experiment like the one presented in this paper, we changed the definition of the budget, dimension, and scaling (budget $25,50,100,200,400,800$, scaling-factor $1$, and add a limit $40$ to the dimension) in the \url{ng_full_gym} experiment in \url{nevergrad/benchmarks/gymexperiments.py} (Line 80) for obtaining the setup as in Section III of the main paper.

\section{Bigger neural nets for OpenAI Gym}
\label{os0}
Here, we offer additional details about the configuration employed to produce the results depicted in Fig. 5 in the main manuscript. Bigger neural networks provide a more complete version of the Ng-Full-Gym problem, where the neural factor parameter, which scales the size of the neural networks, is equal to 3. This does not change the optimization methods, only the scale of the problems towards a greater dimension. 
Dimensions up to 264 are considered. 
This benchmark is unbounded, the the scale of the algorithms (i.e., the standard deviation of the first samples) is therefore particularly critical and makes a fair comparison difficult.

We note that PSO and HyperOpt are still the best or among the best for each considered budget. BO-based methods (BO and AX) perform well for limited budgets and algorithms specific for RL tasks (NGOpt16RL and SpecialRL) become always more competitive as the budget increases.

\section{Budgets 800, 1600, and 3200}\label{os}
\begin{figure}[h!]\centering
\vspace{-0.4cm}\subfloat[Budget 800, tiny neural net.]{\includegraphics[angle=0,  clip,width=.4\textwidth]{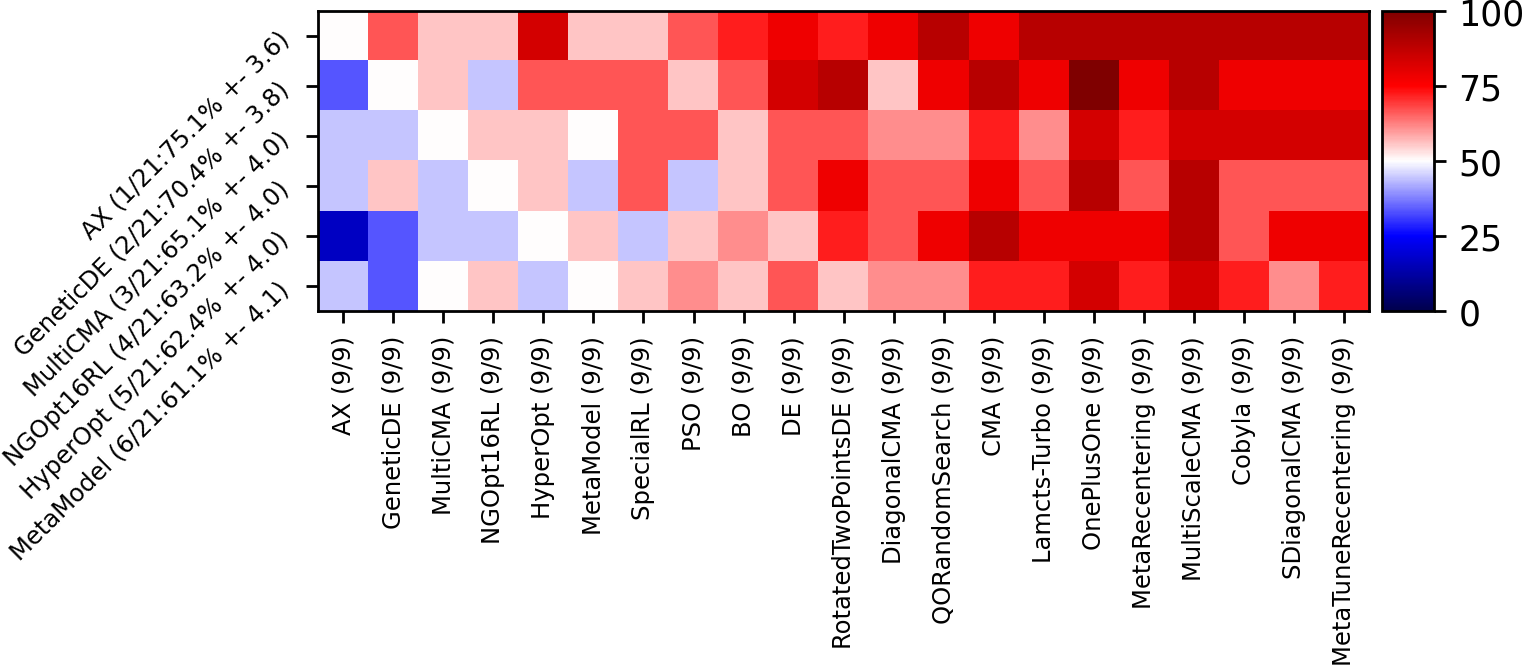}}\\
\vspace{-0.4cm}\subfloat[Budget 800, big neural net.]{\includegraphics[angle=0,  clip,width=.4\textwidth]{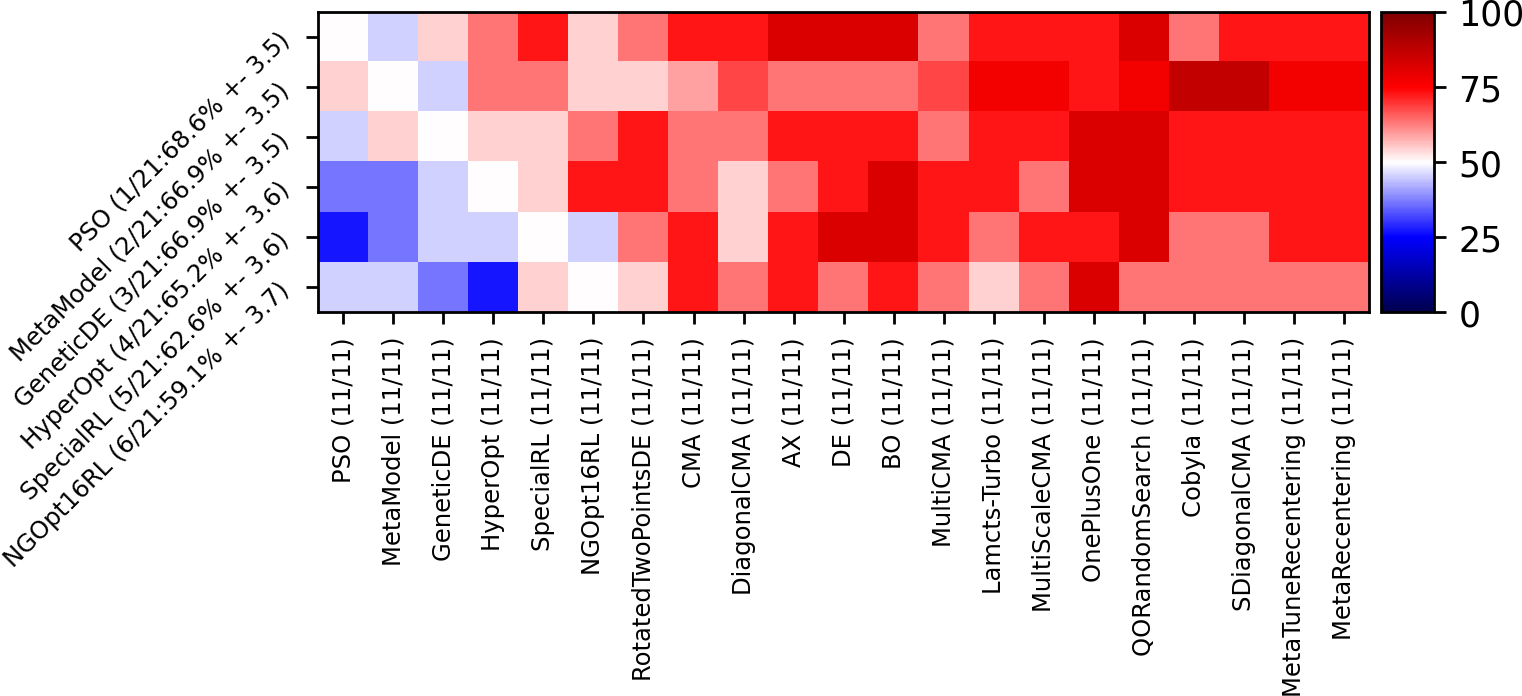}}\\	
\vspace{-0.4cm}\subfloat[Budget 1600, tiny neural net.]{\includegraphics[angle=0,  clip,width=.4\textwidth]{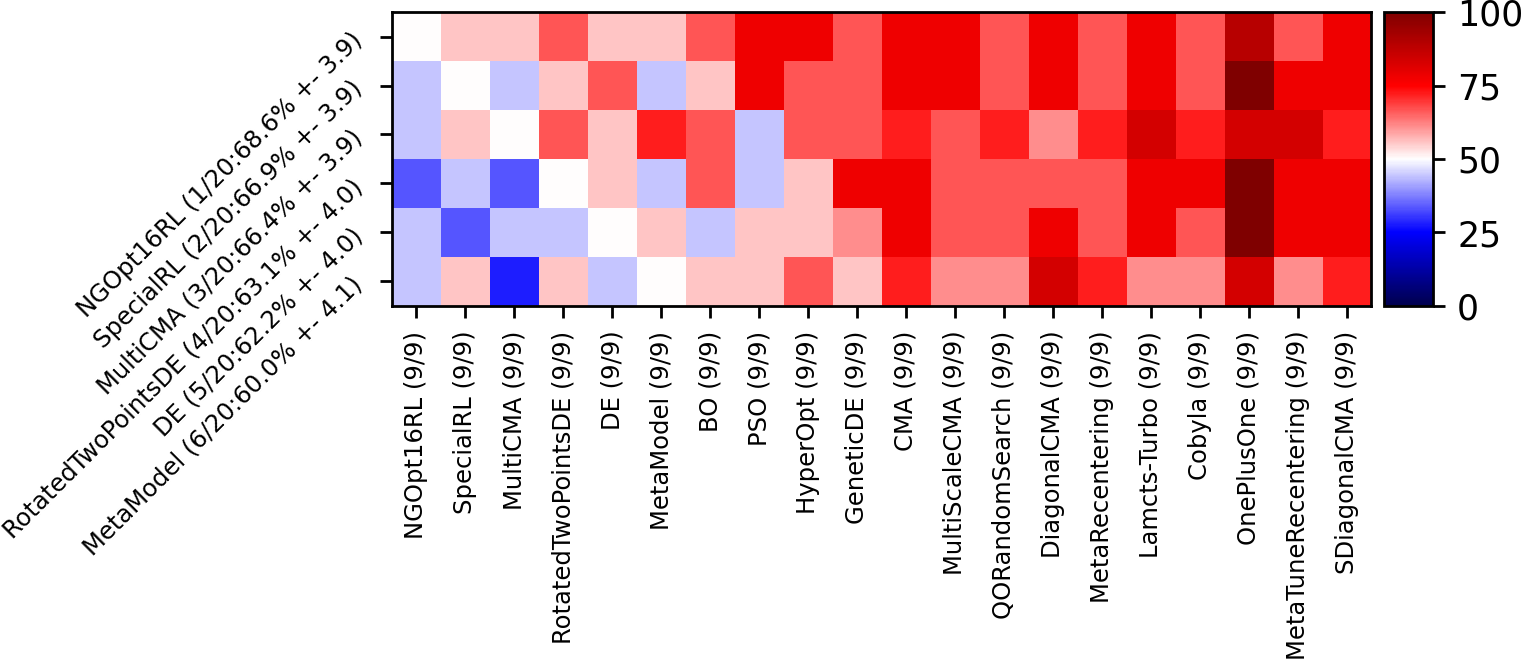}}\\
\vspace{-0.4cm}\subfloat[Budget 1600, big neural net.]{\includegraphics[angle=0,  clip,width=.4\textwidth]{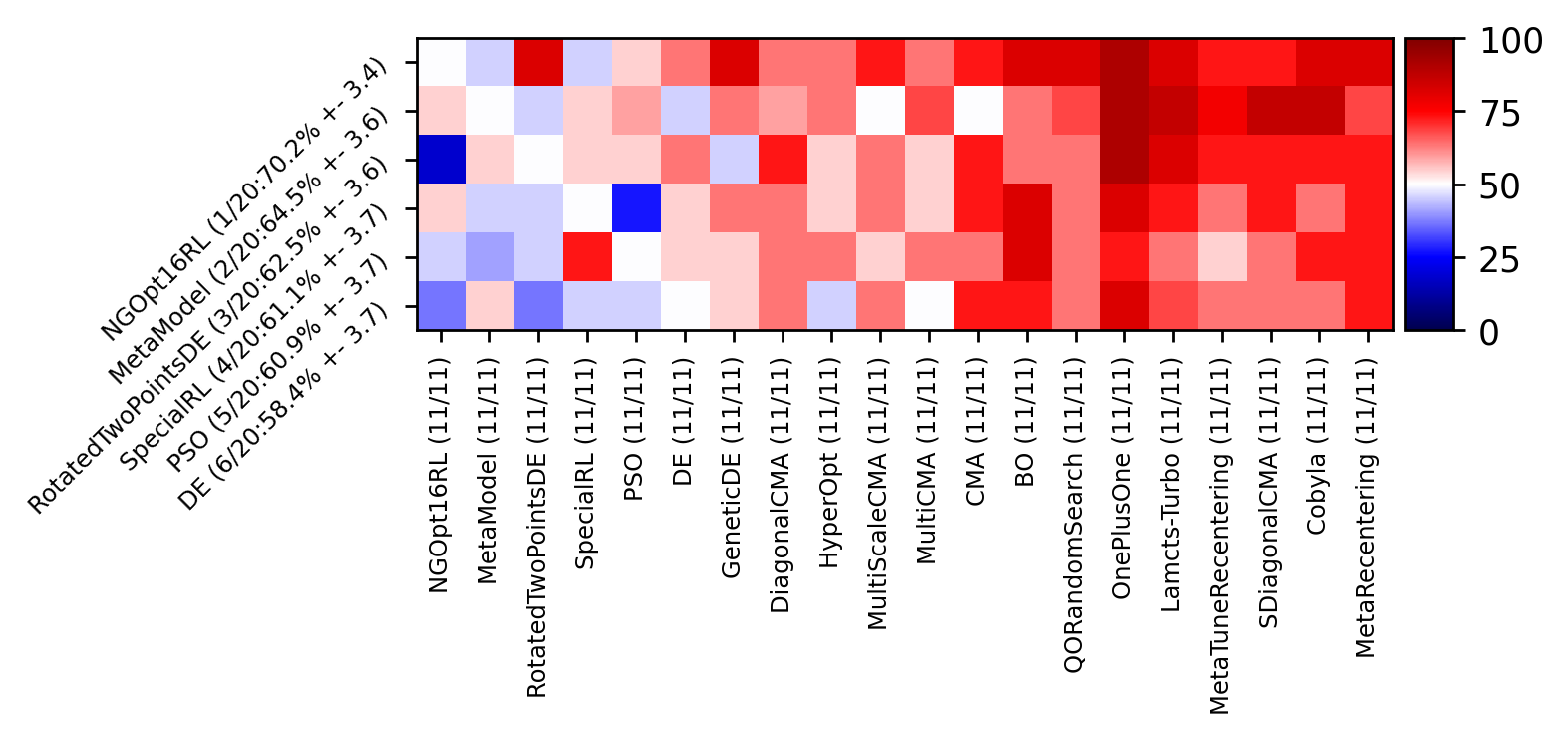}}\\	\vspace{-0.4cm}\subfloat[Budget 3200, tiny neural net.]{\includegraphics[angle=0,  clip,width=.4\textwidth]{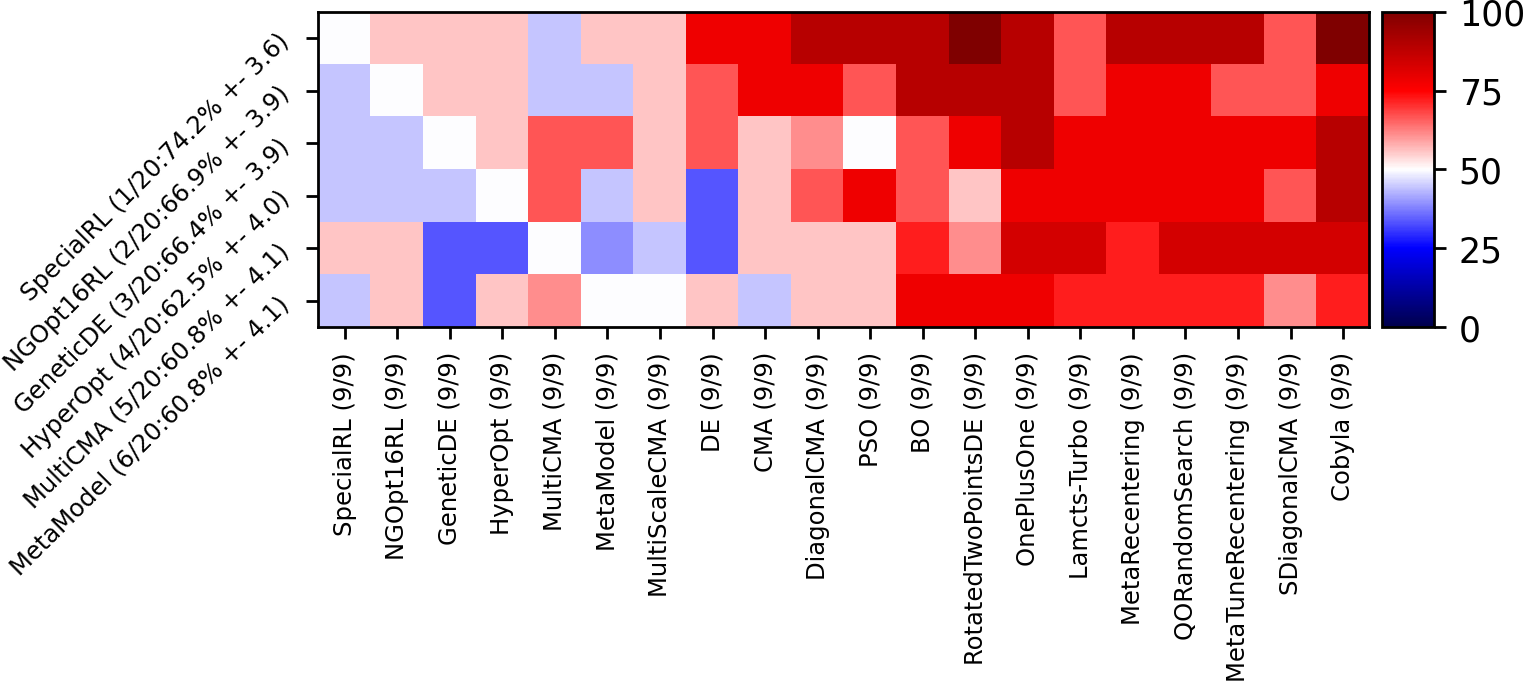}}\\
\vspace{-0.4cm}\subfloat[Budget 3200, big neural net.]{\includegraphics[angle=0,  clip,width=.4\textwidth]{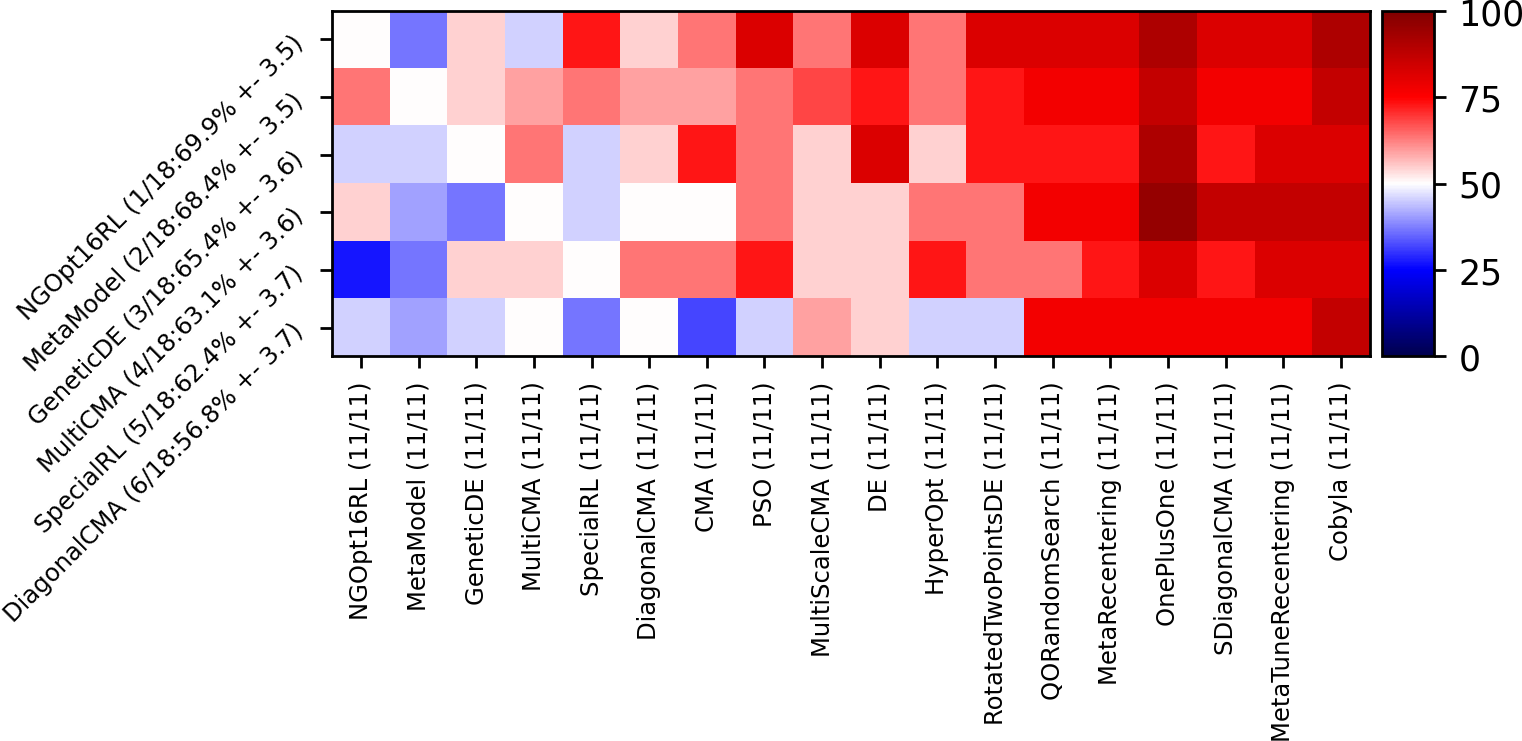}}
\caption{Extension of Figs. 4 and 5 in the main manuscript, for budget 800, 1600, and 3200, for algorithms that were sufficiently fast (faster than 3 days of wall-clock time).\label{det4}}
\end{figure}

While Figs. 4 and 5 in the main paper present results restricted to budgets of at most $400$ function evaluations, Fig.~\ref{det4} presents results for larger budgets of 800, 1600, and 3200 evaluations applied to multi-deterministic Open AI Gym with both tiny and (for methods which are computationally cheap enough) bigger neural nets. 
As in \cite{MeunierRWRRTMD22}, CMA or NGOpt get better as the budget grows. We also find that, while most BO-based methods weaken, HyperOpt performs satisfactorily.

\section{List of methods}\label{algs}
Table \ref{tab3} contains all algorithms involved in our comparison. They are listed in alphabetical order. For each method, we give the labels used in the plots, the testbeds on which they were tested (BBOB, OpenAI Gym, or both), and a brief description of the algorithm.
\begin{table*}
\begin{center}
\caption{Algorithms involved in our comparison. OAIG = OpenAI Gym}
\label{tab3}
\begin{tabular}{| p{.18\textwidth} | p{.133\textwidth} | l | p{.5\textwidth} |}
\hline
\textbf{Name} & \textbf{Label} & \textbf{Testbed} & \textbf{Description}  \\ 
\hline
AX~\cite{ax} & ax, AX & BBOB, OAIG & A modular BO framework that uses BoTorch primitives for optimization over continuous spaces. It automates the selection of optimization routines, reducing the amount of fine-tuning required.\\ 
\hline
BO~\cite{bo} & BO & BBOB, OAIG & The Bayesian Optimization algorithm~\cite{bo} implemented in Nevergrad. The python class is a wrapper over the bayes\_opt package \cite{bopackage}. \\ 
\hline
CMA-ES~\cite{cma} & cmafmin2, CMA & BBOB, OAIG & Covariance Matrix Adaptation Evolution Strategy, a state-of-the-art optimizer in evolutionary computation for non-linear non-convex BBO problems. \\ 
\hline 
Cobyla~\cite{cobyla} & Cobyla & BBOB, OAIG & Constrained Optimization By Linear Approximation optimizer. A sequential trust-region algorithm for derivative-free constrained problems. It employs linear approximations to the objective and constraint functions. \\
\hline
DefaultCMA~\cite{cma} & defaultcma & BBOB & A version of CMA without the BBOB-specific initialization used for the experiments of CMA on BBOB. \\ 
\hline 
Diagonal CMA-ES \cite{cma} & DiagonalCMA & OAIG &   Version of CMA-ES made faster by a diagonal covariance matrix.\\ 
\hline
Differential Evolution \cite{de} & DE & OAIG & Classical differential evolution algorithm.  \\ 
\hline
Genetic DE~\cite{nevergrad}  & GeneticDE & OAIG &  Hybrid algorithm that performs 200 iterations with RotatedTwoPointsDE and then uses TwoPointsDE. \\ 
\hline
HyperOpt~\cite{HyperOpt} & hyperopt, HyperOpt & BBOB, OAIG & Library for serial and parallel hyperparameter optimization, designed to accommodate Bayesian optimization algorithms based on Gaussian processes and regression trees. We use the version based on Parzen estimates.  \\ 
\hline
LA-MCTS~\cite{lamctsNeurips} & lamcts5, Lamcts-Turbo & BBOB, OAIG & MCTS-based derivative-free meta-solver that recursively learns a space partition in a hierarchical manner. Sampling and minimization are then performed in the selected region using TuRBO-1 (no bandit). \\ 
\hline
MetaModel~\cite{nevergrad} & MetaModel & OAIG & Nevergrad's solver that adds a metamodel to an optimizer, which is defined as an input parameter. Unless specified otherwise, the base algorithm is CMA-ES. \\
\hline
MetaRecentering~\cite{CauwetCDLRRTTU20} & MetaRecentering & OAIG & One-shot, fully parallel optimization method using a Hammersley sampling scheme. A variant of Random Search that ensures a higher uniformity. \\
\hline
MetaTuneRecentering \cite{ppsnrescaling} & MetaTuneRecentering & OAIG & Similar to MetaRecentering but with a more sophisticated parametrization. The population is sampled according to a normal distribution with a variance calibrated to the population's size and the problem's dimension.  \\ 
\hline
MultiCMA \cite{nevergrad} & MultiCMA & OAIG &  Nevergrad's solver that splits the search budget in two. In the first portion, it tries different search algorithms; then it uses the rest of the budget on the strategy that worked the best. Here, the compared algorithms are three different initializations of CMA-ES. \\ 
\hline
MultiScale CMA-ES \cite{nevergrad} & MultiScaleCMA & OAIG & Same algorithm as MultiCMA, but the CMA-ES versions have different scales. \\
\hline
NGOpt~\cite{nevergrad} & ngopt16 & BBOB & A wizard that combines many classical algorithms in various ways based on the problem definition, without human intervention~\cite{MeunierRWRRTMD22}. For sequential, low-dimensional, and noise-free problems, it mainly uses CMA, Cobyla, and (1+1)-type sampling equipped with metamodels.\\ 
\hline
NGOptRL~\cite{nevergrad} & NGOpt16RL & OAIG &  NGOpt version optimized for reinforcement learning tasks. It uses a specific bit of information, as it uses the knowledge that the problem is an RL problem. It is a bet-and-run of DiagonalCMA, PSO and GeneticDE.\\ 
\hline
OnePlusOne \cite{oneplusone} & OnePlusOne & OAIG &  (1+1) evolutionary algorithm with the one-fifth adaptation rule. \\ 
\hline
Optuna~\cite{OptunaKDD}& optuna & BBOB & Automatic hyperparameter optimization software framework which uses state-of-the-art algorithms for sampling hyperparameters and pruning unpromising trials. By default, Optuna implements a BO algorithm (Tree-structured Parzen Estimator).  \\ 
\hline
Particle Swarm Optimization \cite{psoreference} & PSO & BBOB, OAIG & 
An algorithm based on moving a population of particles in the search space according to their current positions and velocities.
\\ 
\hline
Quasi Opposite Random Search (inspired by \cite{quasiopposite}) & QORandomSearch & OAIG & Random Search algorithm that symmetrizes exploration with respect to the center but multiplying for a random factor. ``Quasi'' means that the exploration is not exactly symmetric. \\ 
\hline
Rotated Two Points DE\cite{holland} & RotatedTwoPointsDE & OAIG &  Based on TwoPointsDE, which is DE combined with a 2-points crossover. This version can copy and paste from one part of the list of variables to another part: the crossover is not necessarily in place.
  \\ 
\hline
Scaled Diagonal CMA-ES \cite{nevergrad}& SDiagonalCMA & OAIG & Scaled version of DiagonalCMA where the initialization is divided by 1000, i.e., all coordinates are divided by 1000 (if the center is at zero, otherwise translation is applied). \\ 
\hline
SMAC~\cite{smac} & SMAC & BBOB & A sequential model-based algorithm for the hyperparameter optimization of ML algorithms, specifically suitable for high dimensions and discrete input dimensions. \\ 
\hline
SMAC-HPO~\cite{smac} & SMAC2 & BBOB & SMAC with Hyperparameter Optimization. \\ 
\hline
SpecialRL~\cite{nevergrad}  & SpecialRL & OAIG &  Algorithm that runs NGOpt16RL for half the budget, and then uses test-based population size adaptation (TBPSA), an algorithm dedicated to noisy optimization~\cite{vasilfoga}. \\ 
\hline
TuRBO~\cite{turbo} & turbo  & BBOB & A trust-region-inspired algorithm using Thompson sampling rather than the optimization of an acquisition function to find new candidate solutions in each subregion.  \\ 
\hline
TuRBO-20~\cite{turbo} & turbo20  & BBOB &  The multi-trust-regions counterpart of Turbo. \\ 
\hline

\end{tabular}
\end{center}
\end{table*}

\end{document}